# Optimal players policies for discrete and continuous ambush games

Emmanuel Boidot[1] Aude Marzuoli[2] Eric Feron[3] *

***Abstract*—We consider an autonomous navigation problem, whereby a traveler aims at traversing an environment in which an adversary tries to set an ambush. A two players zero sum game is introduced. Players' strategies are computed as random path distributions, a realization of which is the path chosen by the traveler. A parallel is drawn between the discrete problem, where the traveler moves on a network, and the continuous problem, where the traveler moves in the plane. Analytical optimal policies are derived. Using assumptions from the Minimal Cut - Maximal Flow literature, the optimal value of the game is shown to be related to the maximum flow on the environment in both the discrete and the continuous cases, when the reward function for the ambusher is uniform. A linear program is introduced that allows for the computation of optimal policies, compatible with non-uniform reward functions. In order to relax the assumptions for the computation of the players' optimal strategies of the continuous game, a network is created, inspired by recently introduced sampling based motion planning techniques, and the linear program is adapted for continuous constraints.**

## I. Introduction

We seek to develop navigation policies for dynamical systems operating in adversarial environments. An agent tries to operate a vehicle from a given origin, or source $s$, to a desired target set, or sink $t$, while avoiding an undesired set of states (ambushes); another agent sets up ambushes. The traveler is penalized if the vehicle gets caught in an ambush, while the ambusher then gets rewarded. Examples of applications of this research can be found in protection of Cash In Transit (CIT) vehicles.

When the system of trajectories is a finite network, the problem of ambush avoidance can also be described as a network interdiction problem (or NIP), where interdiction are specified on nodes. Of particular interest in this paper is the subclass of maximum flow interdiction problems, where the goal of the interdictor is to prevent the flow of some unwanted items [1]. Maximum flow interdiction problems have been extended to a stochastic formulation to minimize the expected maximum flow [2]. Such problems are usually formulated as sequential zero-sum Stackelberg games, with two steps and two players. Washburn [3] identified optimality properties regarding some interdiction network subclasses. Considering ambush games under the lens of network interdiction problems allows us to better understand the optimal position of the ambushes and predict the optimal outcome of the game. Another well-known framework for path planning in adversarial environments is differential games, pioneered by Isaacs [4]. For a complete and recent set of references, see [5]. A particular type of differential game is the pursuit-evasion game. While these games share some ground with ambush games, the authors want to stress that they do not constitute related or equivalent problems. Unlike pursuit-evasion games, we are interested in the case where one agent is a vehicle, and the other agent makes a unique decision of placing one or more ambushes. Our approach naturally leads to the design of random distributions of trajectories. For the same scenario, the returned strategy is unique, but two vehicles following this strategy might use different paths. For clarity reasons, a probabilistic strategy is referred to as a "route" and a deterministic realization of this strategy as a "path". Ambush games have merit on their own since they represent asymmetric adversarial situations commonly encountered in today's conflicts. Given the topology of the environment, our goal is to address the following questions by means of game theory:

- What are the best policies for the agent and his/her adversary?
- What is the probability that the agent traverses the environment safely?

This work builds upon previous research on game theory by Ruckle [6], [7] who extended Isaac's [4] classical battleship versus bomber duel by working with a two-dimensional environment made of a rectangular array of lattice points [8] and later as a set of tiles [7]. Ruckle identified optimal policies for an agent, BLUE, and his opponent, RED, for different games played on this rectangular environment, with varying conditions on the type of ambush set that RED may occupy or the type of paths BLUE may follow. The main limitations of his work are that it is only applicable to 2D rectangular environments, where no obstacles are present. Ruckle's formulation was extended by Joseph and Feron [9], [10]. They advanced the idea of a variable game outcome at each node of the network. Their formulation was later used in applications for piracy prevention by Vanek et al. [11][12] and cash transport protection by Salani et al. [13]. The work cited so far proposed solutions to the game on a discrete network; however, there has been little analysis regarding the set of optimal solutions. This will be the focus of Section II.

Graph and, more recently, network theory are thoroughly studied topics. There have been numerous applications of Ford and Fulkerson minimal-cut maximal-flow theorem, particularly in the field of Operations Research [14]. However, much less

[1]PhD candidate, Email:eboidot3@gatech.edu.
[2]PhD, Email:amarzuoli3@gatech.edu.
[3]Dutton/Ducoffe Professor, Email:feron@gatech.edu.
*The authors are affiliated with the Guggenheim School of Aerospace Engineering, Georgia Institute of Technology, Atlanta, GA 30332-0150.

Support for this work has been provided by ARO MURI award W911NF-11-1-0046.

interest has been dedicated to the continuous extension of this theory. In this paper, we first recall some key concepts and important results from Strang [15], Mitchell [16] and Polishchuk [17], before applying them to our own work. Making use of the parallel between discrete and continuous flow propagation, this paper proposes to explore the optimal players' strategies for the discrete and continuous ambush game. First, we restate the discrete ambush game problem and show that the optimal value of the discrete game, as measured in terms of probability of agent survival, is a function of the maximal flow capacity of the network, assuming a uniform reward function for the ambusher. Second, assuming that the environment is polygonal, as detailed in Section III, we show that the optimal value of the continuous ambush game depends on the length of the minimal cut (minimal with respect to a certain metric dependent on R, the reaching capacity, or reach, of RED). Third, using the work of Mitchell [16] and Polischuk [17], an optimal strategy of the continuous game can always be found in a polygonal environment $\Gamma$. Moreover, such a strategy can be computed using an algorithm whose time complexity is $O(n \log n + nh)$, where $h$ is the number of holes in the environment $\Omega$, which is a $n$-gon. Fourth, for practical computation of realistic environments, where rich descriptions of the obstacles and boundaries of the environment, such as a polygonal descriptions, are not readily available, we can approximate the optimal policies of both RED and BLUE by solving a linear program corresponding to a similar game played on a network representative of the environment.

Section II describes the discrete ambush game and presents the analytical results regarding the optimal policies for both players. These results are based on an extension to classical network flow propagation theory for networks with constrained flow capacity at vertices, whose proofs are detailed in Appendix VI-A. A linear program is formulated for the resolution of a similar ambush game on a network and the optimal strategies are derived. Section III serves the same purpose regarding the continuous ambush game. In Section IV, we propose to approximate the continuous problem using a network representation where vertices are sampled uniformly over the environment. Section IV shows that, as the number of vertices goes to infinity, the optimal value of the game played on a sampled network converge to a value shown to be the optimal value of the game for the corresponding environment, as discussed in Section III. This convergence is proven for a special category of networks created through asymptotically optimal graph-based path planning algorithms. Section V discusses the results obtained and concludes the paper with a possible extension of this work.

## II. Ambush Games on Networks

### A. Discrete Game Description

The problem of interest is to plan a path for an agent that needs to move from an origin, or source $s$, to a destination, or sink $t$, in an environment where hostile forces might try to ambush it. The agent may be evolving in the air, on the ground, or at sea. The problem is modeled as a two-player non-cooperative zero-sum game, meaning that a player's sole goal is to optimize his/her personal gain and the sum of the outcome over all players is zero. Player 1, denoted BLUE, chooses a route from origin to destination. Player 2, denoted RED, selects a number of locations (*ambush areas* or *ambush sites*) where he/she can set up one or several ambushes. The number of locations chosen by RED is dependent on its ambushing capability. Ambush locations are chosen from a fixed set of potential locations known to both RED and BLUE. An example of such a decision flow is represented in Figure 1. The number of ambushes depends on RED's resources, and constitutes a parameter of the game.

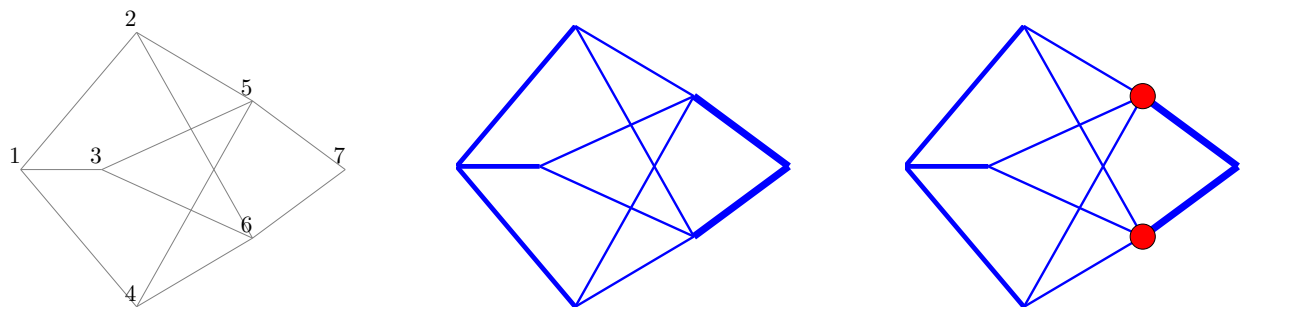

Fig. 1. Decision flow - The network representation of the environment (left) is considered in order to create a strategy for BLUE (middle): $\mathbf{p} = (p_{12}, p_{13}, p_{14}, p_{25}, p_{26}, p_{35}, p_{36}, p_{45}, p_{46}, p_{57}, p_{67}) = (1/3, 1/3, 1/3, 1/6, 1/6, 1/6, 1/6, 1/6, 1/6, 1/2, 1/2)$. Then a strategy for RED (right) is computed: $\mathbf{q} = (q_1, \ldots, q_7) = (0, 0, 0, 0, 1/2, 1/2, 0)$. The outcome of the game for these strategies is $\mathcal{V} = 1/2$. The probability associated with each edge is displayed as the width of the edge.

If BLUE's path passes through an ambush site, then RED wins. If BLUE's path avoids all ambushes, then BLUE wins. The outcome of an ambush corresponds to the casualties that the agent would experience, should an ambush be effectively be present at that location. It is dependent on the characteristics of the local environment. The outcome of a game is the sum of this local outcome over all the ambushes that BLUE's path has gone through. Thus we assume ambushes are not lethal, although high costs may be incurred. The outcome $\alpha(x)$ of an ambush at any point $x$ is assumed to be known by both players. In the remainder of this section we assume that RED only sets one ambush. However RED is allowed to have a mixed strategy, therefore all ambush sites are considered.

The environment is represented by a network $(V, E)$, where $V$ is the set of vertices and $E$ the set of edges, and a reward function, also denoted as the local outcome function. The set $\{\alpha_k\}$ is the projection of the reward function on $V$. Each ambush area $k$ is associated with a local reward $\alpha_k$. In this paper, only single-stage ambush games are considered, meaning that both players need to decide their strategy at the beginning of the game, and they cannot change it during the execution of the route. The environment and its discrete representation are the only information shared by RED and BLUE: the network, the local reward function, BLUE's origin and destination points and the potential position of the ambushes are known to both players, but not the position of those chosen by RED.

A possible strategy for BLUE is represented by a probability vector $\mathbf{p}$, where $p_{ij}$ is the probability that the agent uses edge $(i, j)$ between vertices $v_i$ and $v_j$. Similarly, a strategy for RED is represented by a probability vector $\mathbf{q}$ that contains the probability $q_k$ that RED set an ambush at the ambush location $a_k$. Given the network and the reward function, the goal of the game is to find a joint optimal strategy for BLUE, $\mathbf{p}^*$, and RED, $\mathbf{q}^*$. Consider the case of a recurring transition from state $s$ to state $t$, where there exist two distinct paths from $s$ to $t$. A deterministic approach will return only one of the two paths, whereas the mixed strategy $p$ will return a probability $p_1$ of BLUE using the first path and a probability $p_2$ of BLUE using the second path. Incorporating mixed strategy solutions means that, for a given environment, RED can at best (after a large number of runs following this strategy) figure out the strategy $\mathbf{p}$ but will never gather any certainty over the agent's exact path (or realization of the strategy).

### *B. Mathematical Formulation*

Let $G = (V, E)$ be a network with source $s$ and sink $t$. Consider the case of ambushes being set at vertices: ambush locations $\{a_i\}_i$ and vertices $V = \{v_i\}_i$ are represented by the same set. The case where ambushes are set on a group of vertices belonging to a common geographic area will be discussed in Section IV.

**Definition II.1.** *(Player's Strategy)* A strategy for player BLUE is a mapping $p$ from $E$ to $[0,1]$ such that $0 \leq p(i,j) \leq 1$ and the flow constraints are satisfied, where $i$ and $j$ are vertex indices. A strategy for RED is a mapping $q$ from $V$ to $[0,1]$ such that $0 \leq q(j) \leq 1$ and $\sum_{j \in Q} q_j = 1$. BLUE's (resp. RED's) strategy space $\mathcal{P} = [01]^E$ (resp. $\mathcal{Q} = [01]^V$) is the set of all of these mappings.

From here on, $\mathbf{p}$ is the vector representation of the image of $E$ by the mapping $p$ and $\mathbf{q}$ will be the image of $V$ by the mapping $q$. The probability that BLUE use edge $e_{ij}$ is denoted by $p_{ij}$. The probability that RED set up an ambush at vertex $v_j$ is denoted by $q_j$.

Assume that the two players' strategies are independent. At each vertex $v_j$, the probability that BLUE be ambushed is equal to the probability that BLUE's path go through $v_j$ times the probability that RED set an ambush at this vertex. The gain for RED at this vertex being $\alpha_j$, the expected outcome of the game relative to this vertex is: $\sum_{i|(i,j) \in E} p_{ij} q_j \alpha_j$. Therefore the outcome of the game is :

$$\mathcal{V} = \sum_{j \in V} \sum_{i|(i,j) \in E} p_{ij} q_j \alpha_j = \mathbf{q}^t \mathbf{D} \mathbf{p}. \tag{1}$$

with $D_{jk} = \alpha_j$ if the $k^{th}$ line of $\mathbf{p}$ represents the probability that BLUE use an edge $e_{ij}$ directed towards $n_j$, and $D_{jk} = 0$ otherwise.

The objective of the approach is to find a strategy for BLUE that minimizes the largest possible outcome for RED.

Provided that $q_j \leq 1$ for all $j$, RED can always maximize $\mathcal{V}$ by choosing the vertex $v_j$ for which the probability of BLUE passing through that vertex, weighted by the value $\alpha_j$, is maximal. Therefore BLUE's optimal solution is to minimize this product across all vertices:

$$\mathbf{p}^* = \arg\min_{\mathbf{p}} \left( \max_{j \in V} \sum_{i|(i,j) \in E} p_{ij} \alpha_j \right). \tag{2}$$

Additional constraints enforce the conservation of the flow of probabilities through the network. The probability that the agent arrives at vertex $v_j$ is equal to the probability that the agent leaves the same vertex. Probabilities of the agent going through origin and destination vertices are equal to 1.

$$\begin{cases} \sum_{i|(i,j) \in E} p_{ij} = \sum_{k|(j,k) \in E} p_{jk}, \quad \forall j \in N \setminus \{s,t\} \\ \sum_{j|(s,j) \in E} p_{sj} = 1 \\ \sum_{j|(j,t) \in E} p_{jt} = 1 \end{cases} \tag{3}$$

### *C. Optimal strategies on networks with uniform reward function*

The goal of the optimization problem introduced in (2) is to design a flow of unitary capacity on a graph $G$ from a vertex $s$ to a vertex $t$ that minimizes a given cost function. Concepts of flow propagation on graphs with vertex flow capacity constraints are introduced in order to provide necessary material for the study of optimal solutions to (2). The idea of vertex flow capacity constraint being different from the usual flow propagation setup, these concepts require some graph theory proofs to be adapted to this type of graphs.

**Definition II.2.** [18] *(Vertex Cut)* A vertex cut $K$ in $G$ is a bundling of $N$ of the form $[S, C, S]$, where $s \in S$ and $t \in \overline{S}$ : Such a cut is said to separate $s$ and $t$. The capacity of a vertex cut $K$ is the sum of the capacities of the vertices in $C$. We denote the capacity of $K$ by $cap\ K$.

*(Vertex connectivity)* For a nontrivial connected graph $G$ having a pair of nonadjacent vertices, the minimum $k$ for which there exists a $k$-vertex cut is called the vertex connectivity or simply the connectivity of $G$ it is denoted by $\kappa(G)$, or $\kappa$. The minimum $k$ for which there exists a $k$-vertex $s$-$t$ cut is denoted $\kappa_{st}$.

**Definition II.3.** *(Minimal Cut)* Let $\mathcal{K}$ be the set of all $s$-$t$ cuts of $G$. A minimal $s$-$t$ cut $K^*$ is a cut such that the total capacity of the cut is minimized:

$$K^* = \arg\min_{[S,C,\overline{S}] \in \mathcal{K}} \sum_{v_i \in C} c(v_i),$$

where $c(v_i)$ is the capacity of vertex $v_i$, i.e. the maximum flow that can enter and exit this vertex. If the capacity of all vertices is uniform over $G$, the minimal cut capacity is equal to the vertex connectivity $\kappa_{st}(G)$, up to a constant factor.

The reader should understand that our definition of a network differs slightly from the traditional definition in network theory because of the limited flow capacity at each vertex. We are interested in finding the maximal number of disjoints paths on $G$, which is related to the maximal flow of the graph when vertex capacity is equal to one. This translates into the fact that only one path per vertex is admissible. Extensions of classic network flow theory to vertex constrained network are stated next. Proofs are detailed in Appendix VI-A. We first prove Proposition II.1, which gives the relation between the value of a flow and the capacity of a cut in a network.

**Proposition II.1.** *In any network $G = (V, E)$ with constrained vertex capacity and unconstrained arc capacity, the value of*

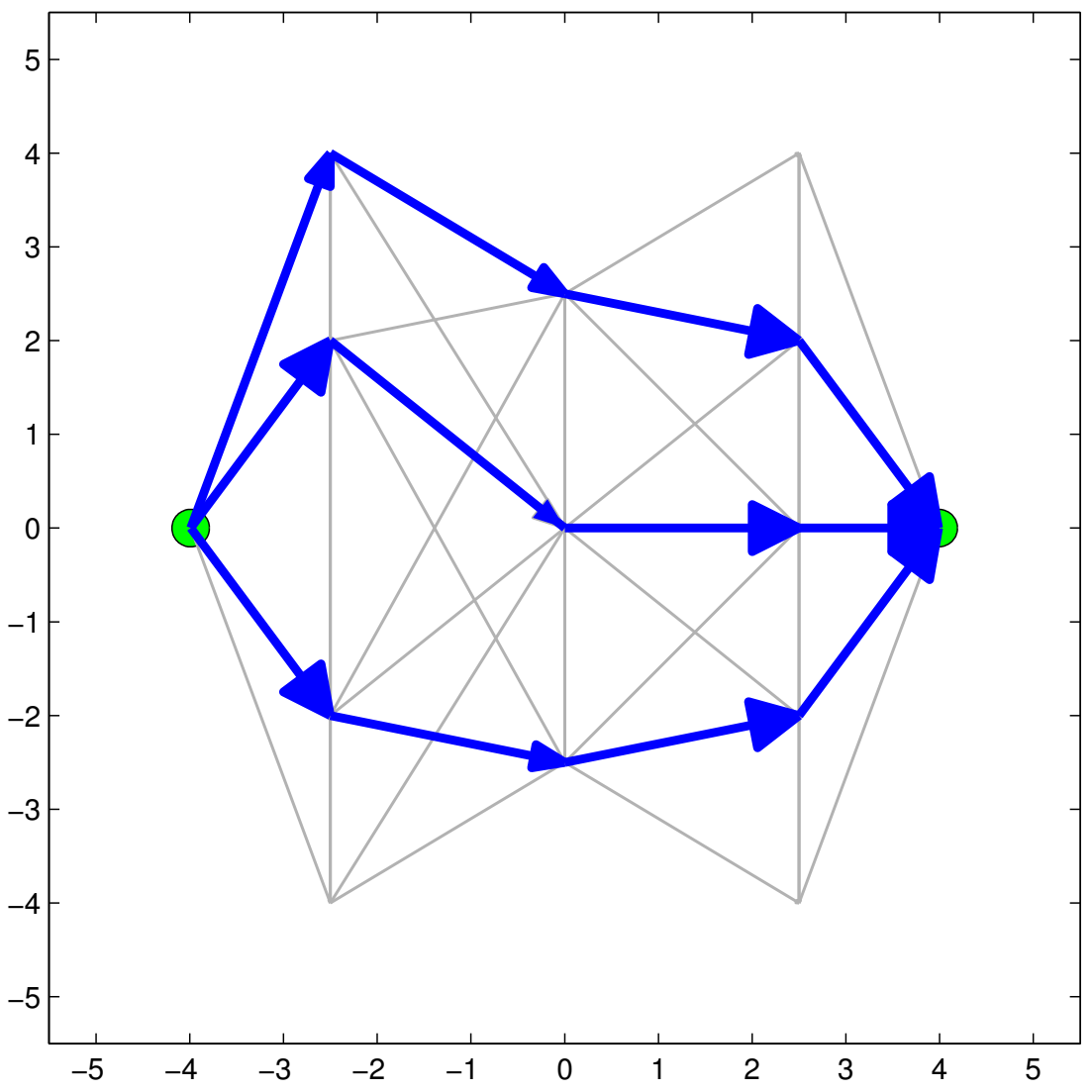

Fig. 2. Number of Disjoint Paths on a graph - The number $\kappa$ of vertex-disjoint paths on a graph where all edges and nodes have unitary capacity is equal to the capacity of the vertex minimal cut. Here $\kappa = 3$.

*any flow $f$ is less than or equal to the capacity of any vertex cut $K$.*

We now extend Ford-Fulkerson Theorem, equating minimal cut to maximal flow, to networks with limited vertex capacity.

**Theorem II.1.** (Constrained vertex capacity Min-Cut Max-Flow) *In a given network $G$ (with source $s$ and sink $t$, and constrained vertex flow capacity), the maximum value of a flow is equal to the minimum value of the capacities of all the vertex cuts in $G$.*

Recall that, for a network with unitary arc flow capacity, Menger's theorem equals the maximal flow and the number of arc-distinct paths on a graph. Menger's theorem has two limitations regarding our problem: it does not constrain flows on vertices and it considers only arc-disjoint ($s$-$t$)-paths. The following theorem relates the min-cut capacity and the maximum number of vertex-disjoint paths on $G$ in the case flow capacity constrained vertices. In the remainder of this paper, a set of disjoint ($s$-$t$)-paths on $G$ denotes a set of vertex-disjoints paths (except in $s$ and $t$) from $s$ to $t$.

**Theorem II.2.** (Menger's theorem for vertex-constrained networks) *Let $G$ be a network with source $s$ and sink $t$. Let each arc and each vertex of $G$ have unit flow capacity. Then, the value of a maximum flow in $G$ is equal to the maximum number $\kappa$ of vertex-disjoint directed ($s$-$t$)-paths in $G$.*

From here on, we consider $G = (V, E)$ to be a network with source $s$ and sink $t$ and unitary vertex capacity everywhere. The term min-cut will refer to the vertex ($s$-$t$) min-cut of $G$. We are interested in identifying properties of the optimal solution related to networks, where the cost of an ambush is the same at each vertex and is equal to 1. Throughout this analysis, assume the capacity of each vertex is 1 and consider an ambush game on a network with a uniform reward function: $\alpha(v) = 1,\ \forall v \in V$. Given this unitary reward function, the outcome of the game can be understood as the probability that BLUE be ambushed by RED. We use Theorem II.2 to define an equidistributed strategy for both players.

**Definition II.4.** *(Flow Equidistributed strategy for BLUE)* Assume $G$ has a min-cut $[S, C, \overline{S}]$ with capacity $\kappa$. Let $\{f_k\}_{k=1}^{\kappa}$ be a set of $\kappa$ disjoints ($s$-$t$)-paths. The flow equidistributed strategy for BLUE is the strategy such that $p_{ij} = \frac{1}{\kappa}$ for all edges $(i, j) \in E$ such that $(i, j) \in f_k$ for some $k$ and zero elsewhere.

*(Equidistributed strategy for RED)* Assume $G$ has a min-cut $[S, C, \overline{S}]$ with capacity $\kappa$. The equidistributed strategy for RED is the strategy such that $q_j = q(v_j) = \frac{1}{\kappa}$ for all vertices $v_j$ in $C$ and zero elsewhere.

**Theorem II.3.** (Game optimal strategy) *Let $\kappa$ be the capacity of a min-cut $[S, C, \overline{S}]$ over $G$. The pair of equidistributed strategies for RED and BLUE constitutes a saddle point, which is also a Nash equilibrium, of the discrete ambush game. The optimal outcome of the game is $\mathcal{V}^* = \frac{1}{\kappa}$.*

*Proof.* Assume that each player's strategy is equidistributed. Note that BLUE's strategy defines an ($s$-$t$)-flow of capacity 1 on $G$ and, for any vertex $j$ in $V$, the flow received by $v_j$ is defined as

$$f^-(v_j) = \sum_{i|(i,j)\in E} p_{ij}.$$

Let $V_f$ be the set of vertices of $G$ with non-zero flow. We have

$$f^-(v_j) = \begin{cases} 1/\kappa, & \text{if } v_j \in V_f \\ 0, & \text{elsewhere.} \end{cases}$$

By definition of a vertex cut, BLUE enters $C$ with probability equal to 1:

$$\sum_{j\in C} \sum_{i|(i,j)\in E} p_{ij} = \sum_{j\in C} \frac{1}{\kappa} = \kappa \cdot \frac{1}{\kappa} = 1.$$

Then, the outcome of the game is:

$$\begin{aligned} \mathcal{V} &= \sum_{j\in V} \sum_{i|(i,j)\in E} p_{ij} q_j, \\ &= \sum_{j\in C} \sum_{i|(i,j)\in E} p_{ij} q_j + \sum_{j\notin C} \sum_{i|(i,j)\in E} p_{ij} q_j, \\ &= \frac{1}{\kappa} \sum_{j\in C} \sum_{i|(i,j)\in E} p_{ij} + 0, \\ &= \frac{1}{\kappa}. \end{aligned}$$

Now, we aim to show that this pair of strategies constitutes a Nash equilibrium. First, assume that BLUE's strategy is an equidistributed strategy and study a variation from the equidistributed strategy for RED.

Consider the case where the probability of ambush is non-zero only on vertices of $V_f$.

$$\begin{cases} q_j \geq 0, & \text{if } v_j \in V_f \\ \quad = 0, & \text{elsewhere,} \\ \sum_{j\in V} q_j = 1. & \end{cases}$$

Then,

$$\begin{aligned}
\mathcal{V} &= \sum_{j\in V}\sum_{i|(i,j)\in E} p_{ij}q_j,\\
&= \sum_{j\in V_f}\sum_{i|(i,j)\in E} p_{ij}q_j + \sum_{j\in V\setminus V_f}\sum_{i|(i,j)\in E} p_{ij}q_j,\\
&= \sum_{j\in V_f}\sum_{i|(i,j)\in E} \frac{1}{\kappa}\cdot q_j + \sum_{j\in V\setminus V_f}\sum_{i|(i,j)\in E} 0\cdot q_j,\\
&= \frac{1}{\kappa}\sum_{j\in V_f} q_j\\
&= \frac{1}{\kappa}.
\end{aligned}$$

Therefore there is no gain for RED when changing his strategy inside $V_f$. Now, assume RED decides to assign a non-zero probability of ambush to a vertex outside $V_f$. Then,

$$\sum_{j\in V_f} q_j = 1 - \sum_{j\in V\setminus V_f} q_j < 1.$$

Therefore,

$$\mathcal{V} < \frac{1}{\kappa}.$$

If BLUE's strategy is equidistributed, then there is no strategy for RED that allows him/her to singlehandedly improve his gain. Consider the case where RED's strategy is equidistributed and BLUE's strategy changes. Given the fact that RED's strategy is non-zero on $C$ only, are only of importance to us the changes in the intersection of the probability flow with the vertices of $C$,

$$\begin{aligned}
\mathcal{V} &= \sum_{j\in C}\sum_{i|(i,j)\in E} p_{ij}q_j + \sum_{j\in V\setminus}\sum_{i|(i,j)\in E} p_{ij}q_j,\\
&= \sum_{j\in C}\sum_{i|(i,j)\in E} p_{ij}\frac{1}{\kappa} + \sum_{j\in V\setminus}\sum_{i|(i,j)\in E} p_{ij}\cdot 0,\\
&= \frac{1}{\kappa}\sum_{j\in C}\sum_{i|(i,j)\in E} p_{ij},\\
&= \frac{1}{\kappa}.
\end{aligned}$$

Therefore if RED's strategy is equidistributed, then there is no strategy for BLUE that allows him/her to singlehandedly improve his/her gain. This shows that the pair of equidistributed strategies is a Nash equilibrium of the game. Given that the game is a zero-sum game, this means that the pair of equidistributed strategies is also optimal, and that the optimal outcome of the game is $\mathcal{V}^* = 1/\kappa$. □

Through this analysis, we have extended Ford-Fulkerson theorem and Menger theorem to networks with limited flow capacity at vertices. Using these results, we have shown that, if the local outcome function is uniform, the optimal outcome of the discrete ambush game is equal to $1/\kappa$, where $\kappa$ is the capacity of the minimal vertex cut of $G$ when $G$ has unitary vertex flow capacity. While similar to optimality results on network interdiction problems, this result is singular in two ways. The first difference is that, in network interdiction problems, flow interdiction appears on edges and not on vertices of the graph. The second difference is that, similarly to general network flow propagation theory, network interdiction problems do not account for vertex flow capacity constraints. Although ambush games had been studied in the past, this result is, to the best of the authors belief, the first complete proof relating ambush games and maximal flows on networks.

While existing algorithms for maximal flows computation could have been extended for our specific problem, we propose a different approach, relying on a linear program formulation, that will allow us to compute optimal solutions of the game in cases where there is no analytic solution of the optimal strategies for both players. Examples of such situations are environments with non-uniform local outcome function. This approach is detailed in Section II-D.

### D. Numerical resolution

The optimal problem defined in (2) is solved as a linear program by introducing the slack variable $z$ satisfying:

$$z \;\geq \sum_{i|(i,j)\in E} p_{ij}\alpha_j \quad \forall j\in N \tag{4}$$

Rewriting Equations (2),(3) and (4), the problem can be posed as the linear program

$$\boxed{\begin{array}{rrl} & \text{minimize } z & \\ \text{subject to} & \mathbf{D}\mathbf{p}-\mathbf{1}z & \leq 0,\\ & \mathbf{A}\mathbf{p} & = \mathbf{b},\\ & \mathbf{p} & \geq \mathbf{0},\end{array}} \tag{5}$$

where $A$ and $b$ represent flow conservation constraints. The probability of each edge being used is computed as to minimize expected losses.

A simple example is displayed in Figure 1. The local outcome $\alpha$ is assumed to be equal to one at each internal node of the network and zero on departure and arrival nodes. The result of the optimization problem presented above is displayed in Figure 1(b). The optimization problem was solved using the interior-point algorithm implemented in function `linprog` in MATLAB [19]. Using a symmetric network (with respect to the ($s$-$t$) axis) and a uniform local outcome (constant across the environment), the following observation is made: the probability of the vehicle passing by is spread across the network and preserves the initial symmetry. This property is of particular interest when it comes to avoiding ambushes. It suggests that optimal solutions to the problem are also the most deceptive ones because most paths have similar likelihoods.

## III. Ambush games on polygonal environments

In this section, we tackle the management of ambushes in a continuous environment from a game theoretic standpoint. In the classical ambush model, an autonomous agent (BLUE) must cross an environment where traps have been placed. If the agent comes within a given distance R from the location of the trap, the agent is penalized or suffers a loss. This problem can be seen as a specific case of robotic path planning in the presence of threats, whose formulation in continuous configuration spaces is new to the best of the authors knowledge. Typical applications include the naval battleships versus mines problem [4], where a battleship travels through a canal where bombs have been set up. Other possible applications are: a setting in which an autonomous agent (such as a delivery drone, or a patrol drone) has radio-frequency communications towards a home base with high level goals, but an adversary sets up jamming honey pots in order to take over control of the agent; the routing of an Army Joint Light Tactical Vehicle (JLTV) traveling through a hostile environment.

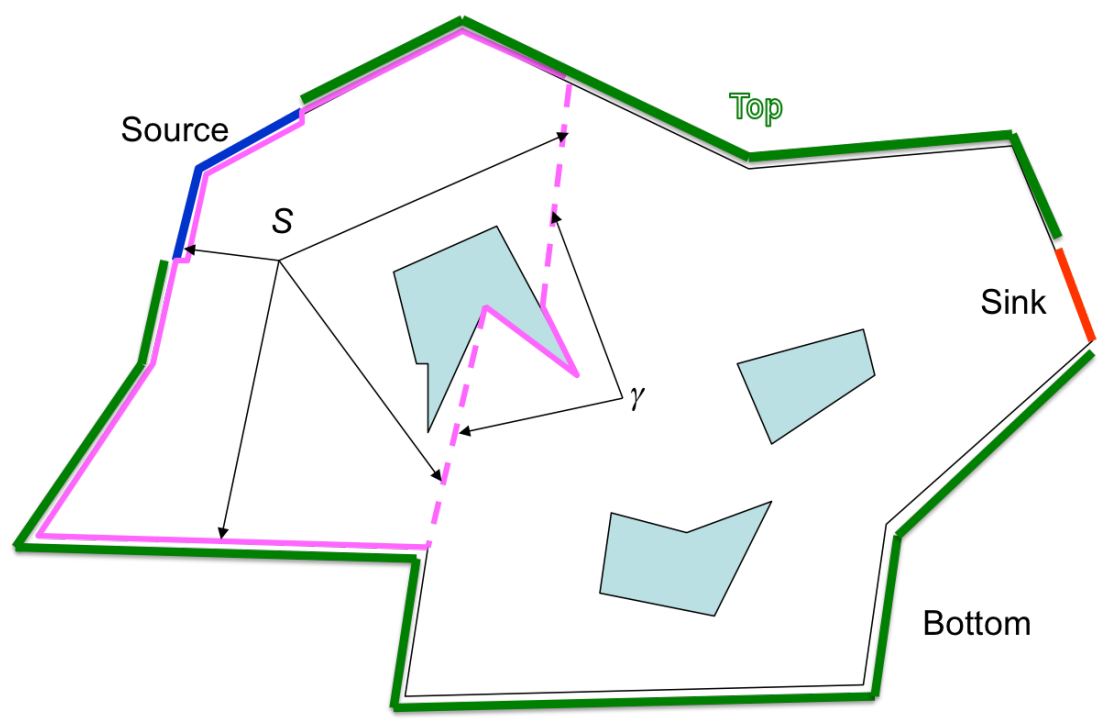

Fig. 3. A cut is the magenta line, labeled $S$. The dashed part of the cut is $\gamma$.

For the remainder of this paper, the environment is supposed to be a polygonal environment, as defined by Strang [15], with the possible presence of holes, also modeled as polygons. This assumption, while strong, remains very realistic for the purpose of studying routing problems in unstructured 2D environments.

### A. Ambush game and minimal cut in polygonal environments

A polygonal domain $\Omega$ is a simple polygon with holes, represented as an ordered set of vertices comprising its outer boundary, and a set of holes, which are also represented as a set of vertices.

Consider a polygonal environment $\Omega$ with a source $s$ and a sink $t$, as displayed in Figure 3. The traveler (BLUE) is trying to go from $s$ to $t$ through $\Omega$, while the adversary (RED) sets up an ambush at some point $x_a$ in the environment $\Omega$, including $\Omega \setminus \mathcal{C}_{\text{free}}(\Omega)$ which represents the part of the environment that the traveler cannot access. The ambusher receives a reward of 1 if BLUE's path comes within a distance $R$ of $x_a$, and receives 0 otherwise. This encounter is a zero sum game and is referred to as a *Continuous Ambush Game in a polygonal environment*. BLUE's path belongs to the set $\mathcal{B}$ of all continuous and piecewise continuous functions $f$ from [0,1] into $\mathcal{C}_{\text{free}}(\Omega)$ such that $f(0)$ belongs to $s$ and $f(1)$ belongs to $t$.

The outcome of the game can be defined as

$$\mathcal{V} =< f, x_a >= \begin{cases} 1, & \text{if } \min\limits_{t\in[0,1]} \|f(t) - x_a\| \leq R \\ 0, & \text{if } \min\limits_{t\in[0,1]} \|f(t) - x_a\| > R \end{cases}$$

For a fixed ambush radius $R$, we are interested in understanding whether the game has an optimal strategy for both players, and how the structure of the environment and the optimal outcome of the game are related. Note that the optimal policies for RED and BLUE might be mixed strategies.

**Definition III.1.** (*Cut*) A cut is defined as a subset $S$ of $\Omega$ such that $S$ contains the sources and no sinks.

(*Minimal Cut*) Let $\partial S$ be the boundary of $S$. We denote $\gamma$ the part of $\partial S$ that separates the source from the sink, and such that each point of $\gamma$ has a neighborhood contained in $\Omega$. A minimal cut is then a cut $S$ where $\gamma$ has minimal length.

Mitchell [16] provides several important results, summarized here. First, a minimal cut always exists, though it is not always unique. Second, a minimal cut can be computed relatively fast.

**Definition III.2.** (*Bottom* and *Top*) The bottom B (resp. top T) of a polygonal environment $\Omega$ is defined to be the portion of the boundary of $\Omega$ between $t$ and $s$ when following $\partial\Omega$ clockwise (resp. counterclockwise), as displayed in Figure 3.

The notion of the critical graph of a domain is central to finding minimal cuts and maximal flows in geometric domains. Figure 4 (left) shows the critical graph of the example environment displayed in Figure 3.

**Definition III.3.** (*Critical Graph*) [20][16] The critical graph of a polygonal environment $\Omega$ is the complete graph $G_c$ such that every hole $H_i$ of $\Omega$ is represented by a vertex of $G_c$, and so are Top and Bottom. The length $|s_n|$ of an edge $s_n = (i, j)$ of $G_c$ is the euclidean distance between the holes $H_i$ and $H_j$.

Mitchell [16] used the critical graph to formulate and prove the "Continuous MaxFlow-MinCut Theorem". Using the critical graph of $\Omega$, a cut can be described as a set $\{s_i\}_{i=1}^{p}$ of edges of the critical graph. A minimal cut $\gamma^*$ of the example environment is a cut of minimal combined edge length, as shown in Figure 4 (left), in red. Polishchuk [17] introduces the concepts of thick path and well separated path.

**Definition III.4.** (*Thick path*) A thick path $(\pi)^R$ of width $R > 0$ is defined as the Minkowski sum of a curve, or *thin path*, $\pi$ in $\mathbb{R}^2$ (the reference path) and a disk of radius $R$:

$$(\pi)^R = \pi \oplus \mathcal{C}_R = \{x + y | x \in \pi,\ y \in \mathcal{C}_R\}.$$

**Definition III.5.** (*Well separated paths*) A set $\Sigma = \sigma_{k\,k=1..n}$ of $R$ well separated (thin) paths (or $R$-WSP) is a set of paths such that no two paths in $\Sigma$ are separated by a distance smaller than $R$ at any point in the domain:

$$\forall j, k \in 1, ..., n, \qquad \min_{s,t\in[0,1]} \|\sigma_k(s) - \sigma_j(t)\| > R.$$

Polishchuk defines a thresholded version of the critical graph of the domain, where the length of each edge $s_i$ is reduced to the closest smaller multiple factor of $2R$, i.e. $\left\lfloor \frac{|s_i|}{2R} \right\rfloor$. This thresholded critical graph is used in the "Discrete Continuous MaxFlow-MinCut Theorem", which states that the maximum number of thick non-crossing paths in $\Omega$ equals the length of the shortest top to bottom (T-B) path in the thresholded critical graph.

### B. Optimal policies of ambush games

We now leverage min-cut problems in continuous environments to derive the solution to the Continuous Ambush

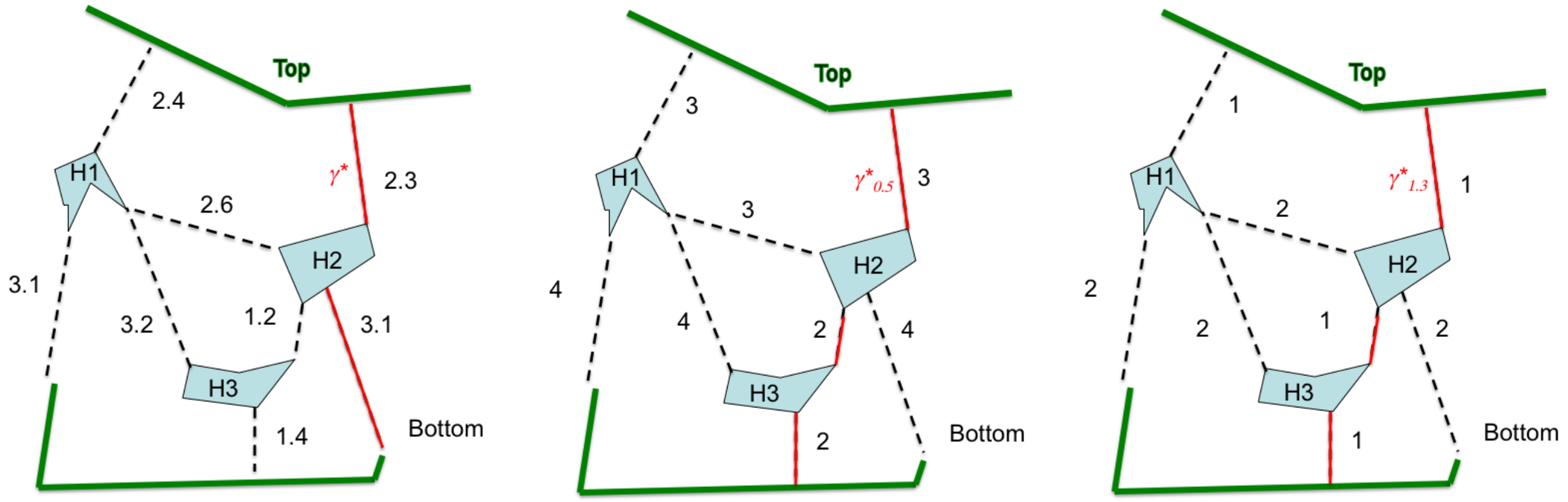

Fig. 4. Example of critical graph and thresholded critical graph for the environment displayed in Fig. 3. (left) Critical graph of $\Omega$ with continuous min-cut in red. (center) Thresholded critical graph for $R = 0.5$. The ambush min-cut is displayed in red and has an ambush-cardinality of $l_{0.5}(\gamma^*_{0.5}) = 7$. (right) Thresholded critical graph for $R = 1.3$. $l_{1.3}(\gamma^*_{1.3}) = 3$.

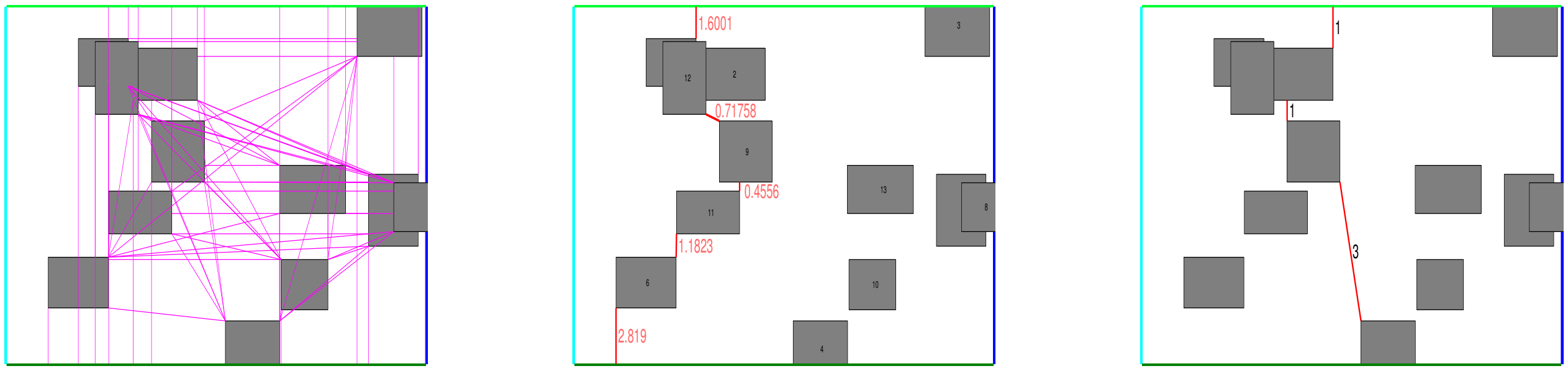

Fig. 5. Example of polygonal environment $\Omega$. The source $s$ is the light blue edge on the left. The sink is the dark blue edge on the right. (left) Critical graph of $\Omega$. (center) Continuous minimal cut as the minimal $T$-$B$ path in the critical graph. (right) Continuous ambush-minimal cut for $R = 1.3$. For this environment $l_{1.3}(\Omega)^* = 5$.

Game in polygonal environments. We introduce the *ambush-cardinality* $l_R(\gamma)$ of a set of segments $\gamma = \{s_i\}_{i=1}^{p}$ as the sum of the rounded up ratio of the length of each segment $s_i$ to $2R$:

$$l_R(\gamma) = \sum_{i=1}^{p} \left\lceil \frac{|s_i|}{2R} \right\rceil .$$

A different thresholded critical graph corresponding to the environment $\Omega$ is defined, where the length of each edge $s_i$ is replaced by $l_R(s_i)$. For the remainder of this section, an ambush-minimal cut $\gamma^*_R$ is supposed to be a cut corresponding to a minimal T-B path in the modified critical graph (upper thresholded critical graph). Let $l_R(\gamma^*_R)^*$ be the capacity of this cut. An example of such a cut is displayed in Figure 4 (center and right) and can be compared with a "traditional" continuous minimal cut displayed in Figure 4 (left). It may be observed that the minimal cut and the ambush-minimal cut or their capacity are not directly related, except in the case where $R$ tends towards infinity and that the ambush minimal cut depends on the value of $R$.

**Theorem III.1.** *The maximum number of $2R$ well separated paths in the domain is equal to the length of the shortest T-B path in the ambush, i.e. $l_R(\Omega)^* = l_R(\gamma^*)$, where $\gamma^*$ is a minimal cut of $\Omega$ with respect to measure $l_R$. Moreover, such a set of $2R$-WSP can be constructed in $O(n \log n + nh)$, where $\Omega$ is a $n$-gon with $h$ holes.*

*Proof.* The existence of such a set of well separated paths for BLUE is discussed by Polishchuk [17]. He states that the number of well separated paths can be significantly larger than the number of thick paths. The number of thick paths can be identified using the uppermost path algorithm described in [17], which runs in time $O(n \log n + nh)$. This theorem is the Discrete Continuous MaxFlow-MinCut theorem (DCMFMC).

Using a reasoning similar to the proof of the Theorem 6.21 [17], we establish Theorem III.1 using the "grass fire" analogy first introduced by Mitchell [16], illustrated in Figure 6. Suppose that the free space $\Omega \setminus H$ is grass over which fire travels at speed 1. Suppose also that the holes are composed of a highly flammable material (the fire burns through a hole at infinite speed) so that as soon as the fire hits a hole, its entire boundary immediately ignites. Ignite the top $T$ at time 0 and initiate the set of $2R$-WSP with the path adjacent to $T$. The wavefront at time $\tau$ is the boundary of the grass that has burnt by $\tau$. The wavefronts make up the streamlines of the flow. The algorithm fills up the free space with the streamlines as the fire propagates until reaching the bottom $B$ so that no more streamlines can be found. The significant events of the algorithm happen when the fire hits an obstacle. Two things happens at an event: the obstacles boundary joins the wavefront, and the streamlines resume their expansion below the obstacle. After burning for $2R$ units of time we can add the wavefront to the set of $2R$-WSP. We then start over, treating the wavefront at $2R$ as new $T$.

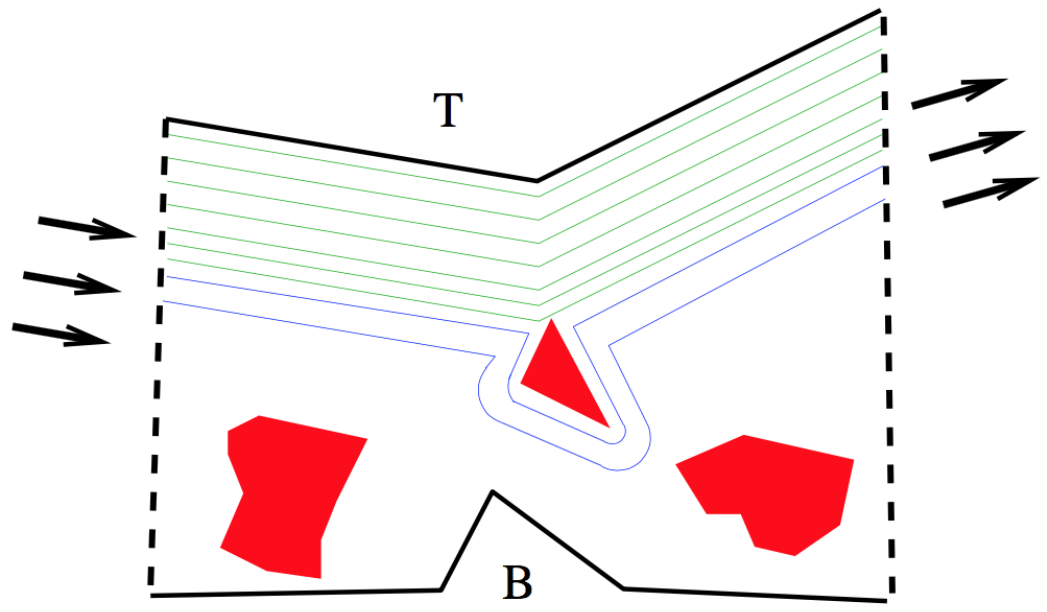

Fig. 6. [17] The wavefronts make up the streamlines of the flow. After hitting an obstacle, the streamlines start going around it.

This algorithm uses the same loop and the same termination condition as the algorithm in the DCMFMC theorem, therefore the time complexity for the WSP computation remains the same as the complexity of the DCMFMC. □

Note that knowledge of the distances between holes and between a hole and Top or Bottom is sufficient to compute the maximum number of $2R$ separated paths. The polygonal assumption on the environment could be relaxed if one were only interested in computing this value. However, the construction of these paths requires this assumption to be enforced.

**Theorem III.2.** *The optimal outcome of the* Continuous Ambush Game in polygonal environments *is* $\frac{1}{l_R(\Omega)^*}$. *An optimal strategy for BLUE is to draw* $l_R(\Omega)^*$ *distinct paths separated by at least* $2R$ *from each other, and to pick each path with probability* $\frac{1}{l_R(\Omega)^*}$. *An optimal strategy for RED is to set* $l_R(\Omega)^*$ *ambushes on an ambush-minimal s-t cut* $\gamma_R^*$ *of* $\Omega$, *where* $\gamma_R^* = \{s_i\}_{i=1}^p$ *is minimal w.r.t. the ambush-cardinality. The optimal placement of ambushes is on segments* $s_i = A_iB_i$ *of* $\gamma_R^*$ *at the following points* $C_{ik}$:

$$C_{ik} = A_i + \frac{2k-1}{2n_i}\overrightarrow{A_iB_i}, \quad \text{for } k = 1, \cdots, n_i; \; i = 1, \cdots, p,$$

*where* $n_i = \left\lfloor \frac{|s_i|}{2R} \right\rfloor + 1$.

*Proof.* Theorem III.1 provides us with the proof of existence of $l_R(\Omega)^*$ $2R$ well separated paths needed for BLUE's strategy. Therefore, assuming that BLUE follows the above described strategy, RED can intercept at most one of these paths. Therefore the probability that BLUE be ambushed is at most $\frac{1}{l_R(\Omega)^*}$.

Next, if RED's strategy is as described, there exists no path from $s$ to $t$ that does not cross $\gamma_R^*$ within a distance $R$ from an ambush point $C_i$. Therefore BLUE is ambushed with probability at least $\frac{1}{l_R(\Omega)^*}$.

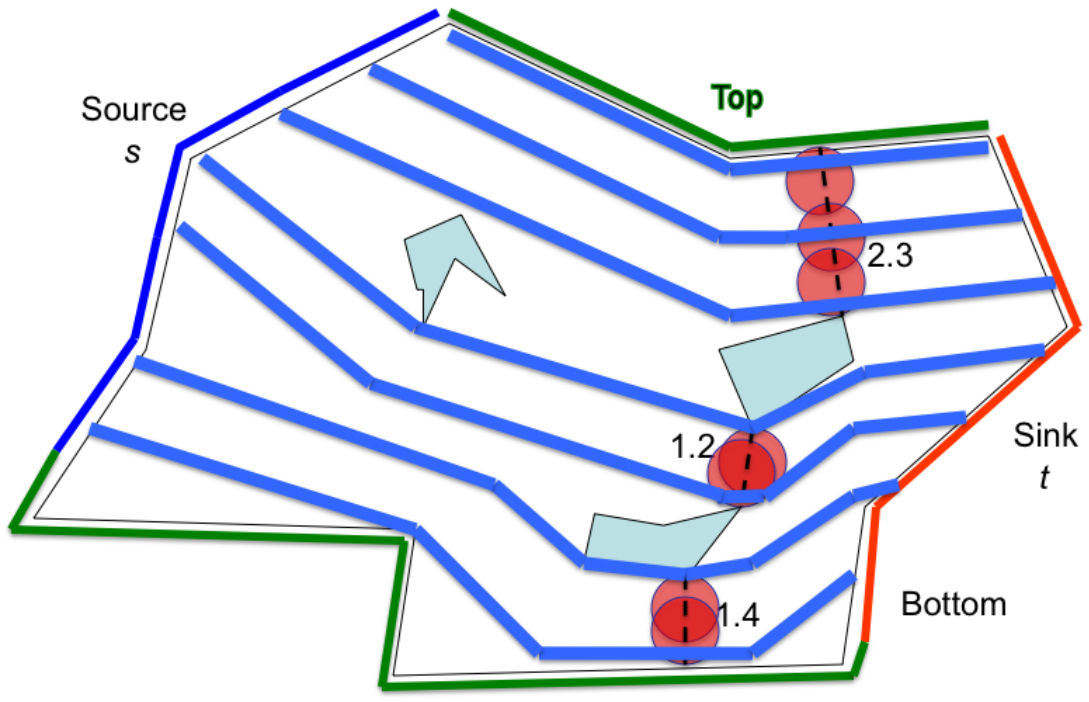

Fig. 7. Example of optimal strategy on a polygonal domain for $R = 1/2$. The dashed line represent the ambush-minimal cut $\gamma_{1/2}^*$, and $l_{1/2}(\gamma_{1/2}^*) = 7$. The number along it represent the length of the corresponding dashed segment divided by $2R$. If a segment's length is $|s_i| = 2.3 \cdot 2R$, it means that it takes at least 3 ambushes to fully intercept the flow going through this segment. The red circles represent RED's strategy. Each one is used with probability $1/7$. The blue lines represent BLUE's strategy. Each path is used with probability $1/7$

Finally, we conclude that the optimal outcome of the continuous ambush game is $\frac{1}{l_R(\Omega)^*}$. Since the strategies described above result in such an outcome, they are optimal. An example of such a strategy for RED and BLUE is displayed in Figure 7, which represents the optimal strategies for RED and BLUE to the Continuous Ambush Game played on the polygonal environment described in Figures 3 and 4. □

Although the existence of BLUE's well separated paths is proven only for polygonal environments, the authors believe that Theorem III.2 could be extended for any bounded environment $\Omega$, where the traveler goes from a source $s$ to a sink $t$, placed on the boundary of $\Omega$ or not. However, the computation of the critical graph, and therefore the computations of the minimal cut and well separated paths that follow on from this graph, become much more strenuous.

In this part, the *Continuous Ambush Game* was introduced and it was proven that the optimal outcome of the game depends on the topology of the environment, through the minimal cut capacity of the thresholded critical graph. Similarly to the reasoning developed in Section II, we could adapt the algorithm developed for the theory of flow propagation in order to compute optimal strategies of the game. However, these method would not be adaptable to non-uniform local reward functions. Therefore a sampling method is developed that allows for the extension of the linear problem formulation to continuous ambush games, as discussed in Section IV.

## IV. Sampled Continuous Ambush Game

### A. Game description

We now propose to solve the ambush avoidance problem numerically through uniform sampling over the environment. The approach developed in Section II relies on a single graph to describe both RED and BLUE's strategies. BLUE's strategy space is the set of all edges in the graph, while RED's strategy space is the set of all nodes. In these setups, ambushes only take place at nodes, therefore RED is unable to intercept flows neighboring his/her ambush even if they are arbitrarily close to his ambush point. Our previous approach [21] used a fixed tiling of the environment in order to take RED's reach radius $R$ into account. However the tiling is origin-dependent and the optimal strategy for both RED and BLUE might vary largely depending on the location of the tiling origin. Moreover, the tile shape cannot be circular, resulting in an undesired non-uniform impact range.

In the formulation below, a parameter $R > 0$ is introduced, as in [21], that defines the radius of the region that RED controls in a single ambush, which we call the reach zone. The parameter $R$ is the same as that in Section III. Working directly with circular reach zones removes the foregoing limitation and makes this discrete optimization problem closer to the continuous problem defined in Section III.

Let $G = (V, E)$ be a graph and let $\mathcal{R} = \{a_k\}_{k \in Q}$ be a finite set of ambush locations covering the environment. The nodes in $V$ and the ambush points are sampled according to a uniform probability distribution over the environment. The topic of graph representation of an environment is a thoroughly explored field and one can imagine using any of the numerous available techniques to create $G$, such as grid-based discretization of the environment of techniques developed for multiple query path planning algorithms, like Probabilistic Roadmaps [22] or Rapidly-Exploring random Graphs [23].

We define the *Sampled Continuous Ambush Game* (SCAG) as follows. The traveler (BLUE) is trying to go from a set of nodes $s$ to a set of nodes $t$ through $G$. He/She can pick any node in $s$ as a starting node and any node in $t$ as a finishing node. The adversary (RED) sets up an ambush at a point $a_k$ in $\mathcal{R}$. The ambusher receives a reward $\alpha_k = \alpha(a_k)$ if BLUE's path comes through a node within a distance $R$ of $a_k$.

Assume that the two players strategies are independent. At each ambush point $a_k$, the probability that BLUE be ambushed is equal to the probability that BLUE's path enter the circle $\mathcal{C}_R(a_k)$, circle of radius $R$ centered at $a_k$, multiplied the probability that RED set an ambush at this point. The expected outcome of the game relative to this ambush point is

$$\sum_{i | e_{il} \in \mathcal{E}_k^R} p_{il} q_k,$$

where $\mathcal{E}_k^R$ is the set of all edges $e_{il}$ in $E$, directed between nodes $v_i$ and $v_l$, such that $v_i \notin \mathcal{C}_R(a_k)$ and $v_l \in \mathcal{C}_R(a_k)$. Therefore the strategic outcome of the game is almost identical to (1):

$$\mathcal{V} = \sum_{k \in \mathcal{R}} \sum_{i | (i,l) \in \mathcal{E}_k^R} p_{il} q_k = \mathbf{q}^t \mathbf{D} \mathbf{p}, \tag{6}$$

with $D_{kj} = \alpha_k$ if $e_{il} \in \mathcal{E}_k^R$, and $D_{kj} = 0$ otherwise, when the $j^{th}$ column of $\mathbf{p}$ represents the probability that BLUE use edge $e_{il}$.

As in the previous games, the problem is to find a strategy for BLUE that minimizes the largest possible outcome for RED. Provided that $q_k \leq 1$ for all $k$, RED can always maximize $\mathcal{V}$ by choosing the point $a_k$ for which the probability of BLUE passing through $\mathcal{C}_R(a_k)$ is maximal. Therefore BLUE's optimal solution is to minimize this product across all nodes :

$$\mathbf{p}^* = \arg\min_{\mathbf{p}} \left( \max_{k \in Q} \sum_{i | (i,l) \in \mathcal{E}_k^R} p_{il} q_k \right). \tag{7}$$

Again, there are additional constraints to enforce the conservation of the flow of probabilities through the network. This problem is solved as a linear program as in (5). The main difference with the approach of Section II is that we sum over the flows entering an area of the environment instead of summing over the flows entering a single node.

A notable feature of the Sampled Continuous Ambush Game is that nothing prevents us from extending it to higher dimensional environments. Indeed, one could imagine creating a graph representation of a $n$-dimensional environment in the exact same way it is done for a two dimensional environment.

The linear program formulation described above requires the existence of a network to optimize the routing policy. Any graph representation of the environment might be suitable for this purpose. In the remained of Section IV below, we are interested in developing incremental sampling approaches for graph creation. In particular, we are interested in observing the influence of the structure of $G$, the number of nodes in $G$ and the number of ambush points on the optimal value of the Sampled Continuous Ambush Game.

### B. *Comparison of analytic and numerical results*

This section compares the optimal outcome obtained through the foregoing approach with the analysis developed in order to check the convergence of the sampling approach to the results obtained in Section III's analysis. This comparison is done on polygonal environments in order to apply existing work on critical graph and minimal cut computation, as discussed in Section III. The linear program described in Equation 5 is implemented on increasingly dense networks in order to understand the influence of the graph density.

For the numerical results discussed in this section, the network was created using two methods. The first method involves deterministic sampling and local connectivity in the form of a an 8-connected grid. The second method comes from the path planning literature. A Rapidly exploring Random Graph is created to represent the environment. The RRG algorithm is detailed in Appendix VI-C. Other network creation methods have been considered and used in prior work [21], such as random sampling with Delaunay triangulation [24] or grid sampling with Delaunay triangulation. In this paper, only the eight-connected grid and RRG representations of the environment are discussed.

Consider first the grid representation of the example environment shown in Figure 5. As shown above, the ambush-minimal cut capacity of this environment for $R = 1.3$ is $l_{1.3}(\gamma^*_{1.3}) = 5$. Therefore the optimal outcome of the Continuous Ambush Game for this environment is $\mathcal{V}^*_{\text{CAG}} = 1/5$. It can be observed in Figure 8 that for smaller, and therefore coarser, networks the full connectivity of the environment is not captured by the network, resulting in a number of $2R$-WSP lower that the theoretic maximum of five. For a network with only twenty-five nodes, there cannot be more than one 2.6-WSP, therefore the optimal outcome of the Sampled Continuous Ambush Game is 1, as seen in Fig. 8 (top left). As the number of vertices in $V$ increases however, more well-separated paths can be found in $G$, and the optimal outcome decreases to $\mathcal{V}_{\text{SCAG}} = 1/3$ first (for $|V| = 81$ nodes, Fig. 8 (top center)) and finally to $\mathcal{V}_{\text{SCAG}} = 1/5$ when the number of nodes in $V$ becomes sufficiently large ($|V| = 900$ nodes, Fig. 8 (top right)). Since the ambush-minimal cut capacity of the environment is 5, there cannot be more than five 2.6-WSP in the environment. Therefore there cannot be more than five 2.6-WSP in the graph either and we can conclude that increasing the grid resolution will not change the optimal value of the Sampled Continuous Ambush Game. These observations are confirmed in Figure 9 (top), representing the optimal outcome of the SCAG as a function of the number of vertices in the network representation of the environment. For the first few iterations of the algorithm (when $G$ has fewer than 25 vertices), there is no $s$-$t$ path on $G$, therefore the outcome is undefined. Then the optimal outcome of the SCAG rapidly decreases from 1.0 to 0.2, which is the optimal outcome of the CAG for the example environment and thus represents a lower bound on the optimal outcome of the SCAG.

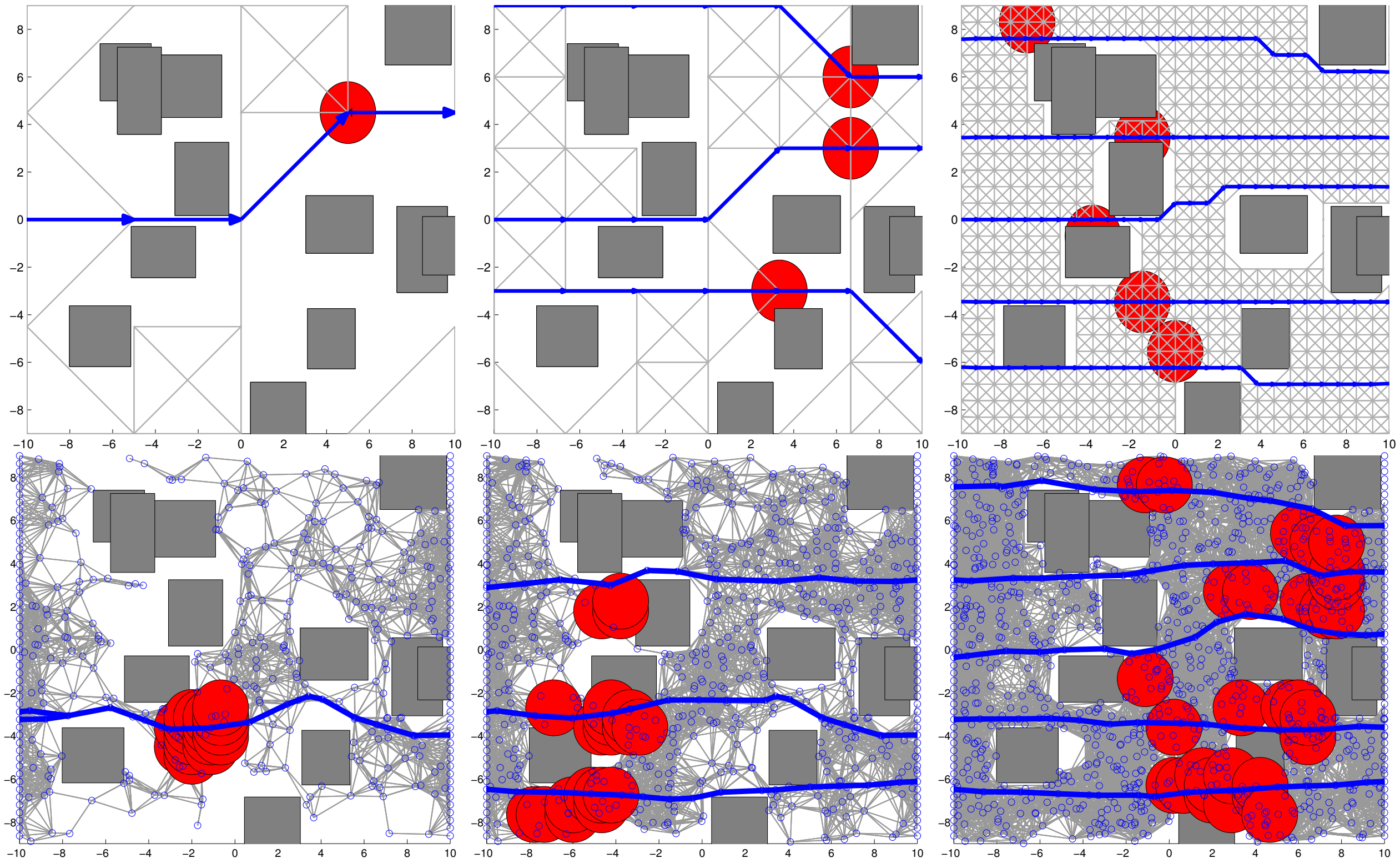

Fig. 8. Optimal policy for RED and BLUE obtained through LP optimization on a sampled network when $R = 1.3$. The upper row corresponds to a grid discretization of the environment with 25 vertices (left), 49 vertices (center) and 729 vertices (right). The lower row corresponds to a RRG discretization of the environment with 200 vertices (left), 500 vertices (center) and 1000 vertices (right). In both cases, the theoretical optimal value of $\mathcal{V}_{\text{CAG}} = \frac{1}{5} = \frac{1}{l_{1.3}(\Omega)^*}$ related to this environment is reached once the graph is dense enough to include 5 well separated paths. Note that there are more ambush areas than $2R$-WSP. This comes from the fact that the linear program is degenerate and there are several optima. For example, a possible class of optima for RED is a strategy such that for any $2R$-WSP $\sigma_k$ on $G$, the sum of the probabilities over all the ambushes that intersect with $\sigma_k$ is $\frac{1}{l_R(\Omega)^*}$.

Looking at Figure 9, we observe that an issue arises from the grid representation of the environment. Although the optimal outcome of the SCAG quickly reaches its lower bound, it can also increase again when the grid density increases. This comes from the fact that the grid from a previous iteration of the algorithm is not conserved in the next iteration. Instead, another deterministic grid is sampled with more vertices, and there is no guarantee that the new grid will conserve as many $2R$-WSP. One way to avoid this problem would be to double the grid density at each iteration of the algorithm, however this would result in very large linear programs after a few iterations, and therefore would lead to much higher computational time per iteration.

A similar analysis can be done for a RRG representation of the environment. Given that the sampling is random, different runs of the SCAG might lead to different optimal outcomes for similar values of the number $|V|$ of vertices in $G$. In order to address this property of SCAG on random graphs, the algorithm is run several times over the example environment shown in Figure 5, and box-plots are used in Figure 9 (bottom) to represent the optimal outcome of the SCAG as a function of $|V|$. Figure 8 (bottom) shows optimal strategies for BLUE and RED for one of these run for different size of $G$. As for the grid representation of the environment, the optimal outcome of the SCAG is lower bounded by the optimal outcome of the CAG for the example environment, which is $\mathcal{V}^*_{\text{CAG}} = 1/5$. While $|V|$ is small, there can be more than one $s$-$t$ path in $G$. As the number of vertices increases, additional $2R$-WSP can be found on $G$, decreasing the optimal outcome of the game to $\mathcal{V}_{\text{SCAG}} = 1/3$ (for $|V| = 600$) and finally $\mathcal{V}_{\text{SCAG}} = 1/5$, which is the optimal outcome of the continuous game for this environment. Figure 9 (bottom) confirms that this convergence can be observed for several runs of the RRG-based SCAG on this environment, thus is not dependent on the specific sampling observed in Figure 8.

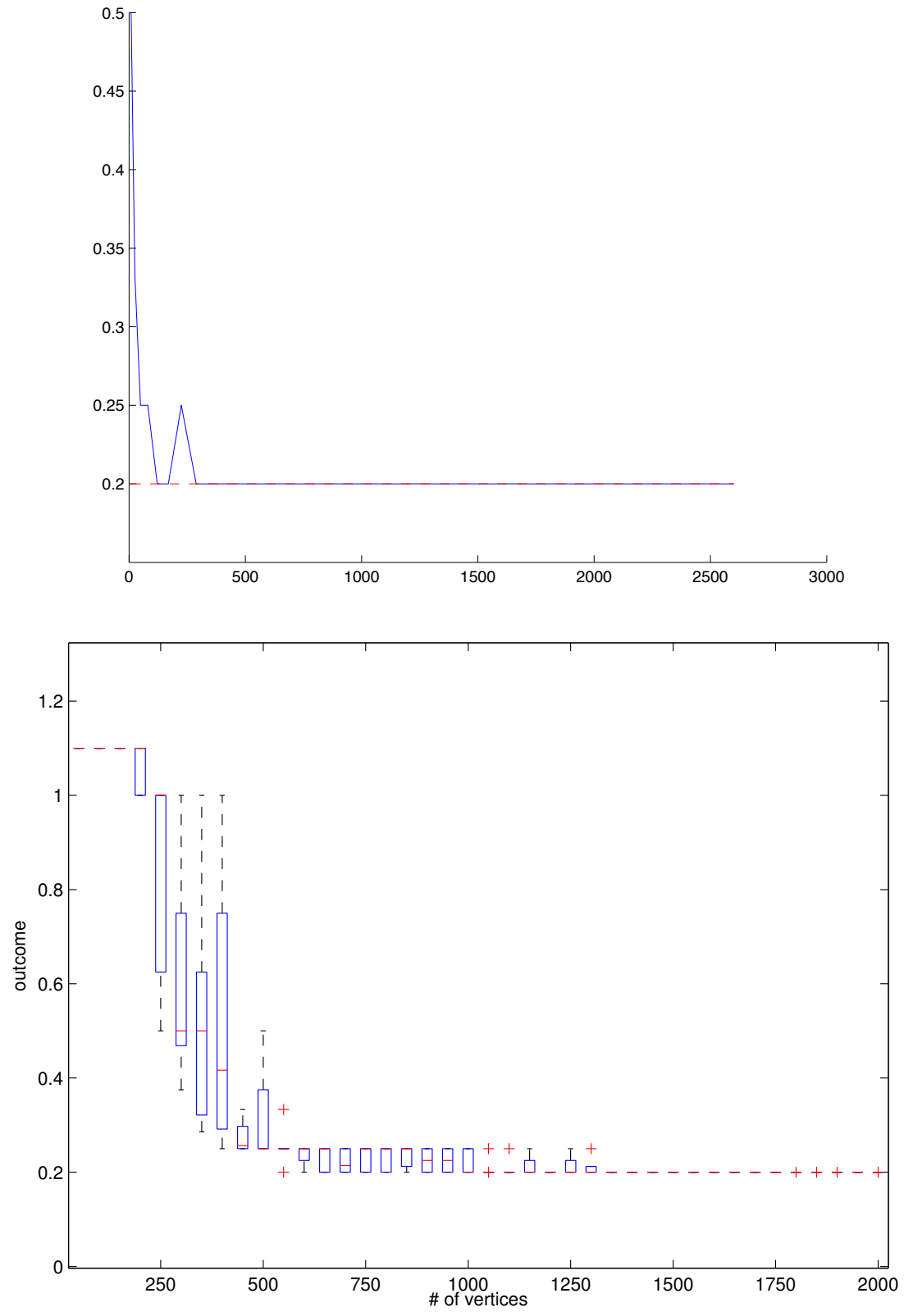

Fig. 9. Optimal outcome of the sampled game as a function of the number of vertices in the network for the environment displayed in Figures 5 and 8. (top) Results for the grid representation of the environment. (bottom) Results for the RRG representation of the environment. Given that the sampling is random, the results are displayed as a box-plot for different cardinalities of network. These boxplots correspond to several runs of the algorithm over the environment shown in Figure 5.

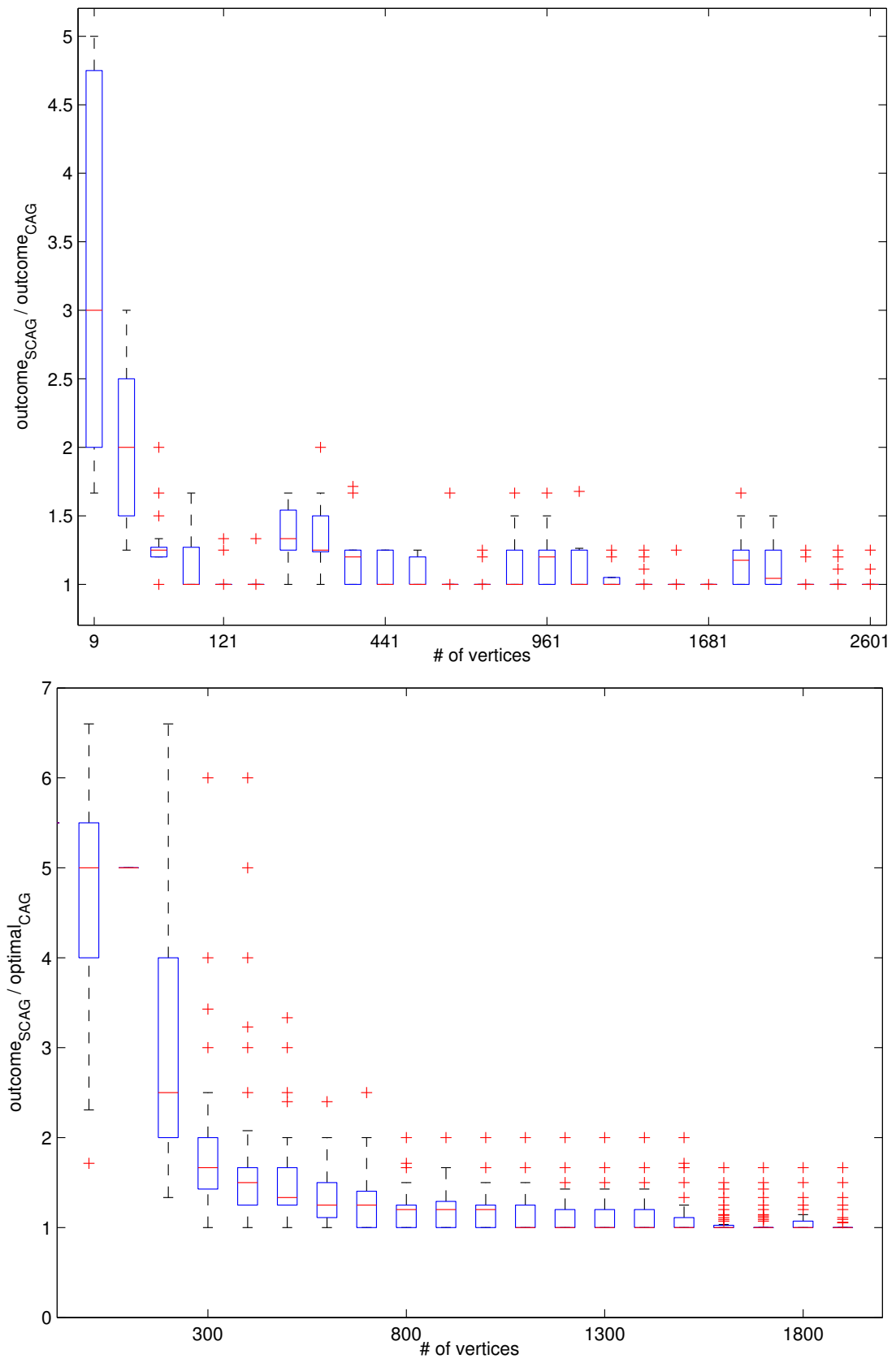

Fig. 10. Ratio of the optimal outcome of the sampled game to the optimal outcome of the CAG as a function of the number of vertices in the graph (averaged over 25 randomly generated environments). In average, the optimal outcome of the sampled games converges to the theoretical optimal value of the continuous game. (top) eight connected grid. (bottom) RRG.
Boxplots are used to display average results and standard deviation over several environments. The standard deviation converges to zero while the mean .

The convergence of the optimal value of the game toward the predicted value for continuous environments has been illustrated in Figure 9 for an example polygonal environment. In order to verify that this convergence could be observed on many polygonal environments, a benchmark was created to compute the optimal outcome of the Sampled Ambush Game on a set of twenty-five randomly generated polygonal environments for different grid resolutions. For each environment, the linear program created from a graph of $|V|$ vertices outputs the value of the optimal outcome, while $|V|$ increases. The outcome of the SCAG as a function of $|V|$ is computed over twenty-five environments. Results are displayed in Figure 10. Again, we observe that the optimal outcome of the Sample Continuous Ambush Game is a decreasing function of the number of vertices in $G$. Additionally, it converges for both methods to the optimal outcome of the Continuous Ambush Game. This value is only dependent on the reach of RED and the structure of the environment, and is equal to the ambush-minimal cut capacity of the environment. As for the example environment, the grid-based SCAG seems to reach the optimal outcome much faster the the RRG-based SCAG. However, we observe again that the outcome is not monotonously decreasing and can increase for some grid resolutions (for example when $|V| = 300$). Additionally, the standard deviation associated with the RRG-based SCAG decreases more steadily than the variance associated with the grid-based SCAG. This means that performances of RRG-based SCAG will be comparable on most environments, whereas there might be a larger variation in grid-based performances.

Looking at the behavior of the SCAG as the number of vertices in $G$ becomes very large, Figure 10 empirically verifies Theorem IV.1.

**Theorem IV.1.** *Assume that the graph supporting the Sample Continuous Ambush game is constructed using a probabilistically complete and asymptotically optimal graph based path planning algorithm. Then the optimal outcome of the Sampled Continuous Ambush Game converges to the optimal outcome of the Continuous Ambush Game as the number of sample points in the environment goes towards infinity.*

Based on Theorem III.2 and Theorem II.3, Theorem IV.1

suggests a relationship between the continuous minimal cut capacity of a polygonal environment and the asymptotic capacity of the cut of a network with increasingly many nodes. Note that having an asymptotically optimal graph based algorithm for the network creation is only a sufficient condition for the optimality of the SCAG. The proof of Theorem IV.1 is given in Appendix VI-C along with some previous results on sampling based path planning. It is only valid for graphs created using PRM$^*$ or RRG algorithms. However, Figure 10 (top) seems to indicate that a similar result exists for decomposition-based methods.

An interesting property of the methods proposed in this section is that they remain valid for applications in a higher dimensional setting, and can be easily extended to aerial vehicles navigation problems for example. This brings up a second issue with grid-based SCAG, expressed in the path planning literature. Given that the goal of SCAG is to approximate the optimal solution of ambush games in the continuous domain, we might be interested in creating graphs in the configuration space ([25]) of the autonomous agent instead of creating them in the geometric space. While the geometric space is limited to low dimensions, in most modern robotic applications, it makes more sense to plan in the configuration space, which can have a much larger number of dimensions. However, decomposition-based methods have been shown to be mostly limited to state spaces with five or fewer dimensions, given that they suffer from the large number of cells in higher dimensions ([26]). This makes decomposition-based methods such as the eight connected grid far less well suited for a large subclass of robotic planning applications. While 8-connected grid appears to be fairly efficient for the 2D environments used in the benchmark, the authors believe that the proven convergence of RRG-based SCAG and their adaptability to higher dimensions makes them a better choice for future applications of SCAG to more realistic autonomous agents models.

## V. Conclusion

The problem of ambush games is of particular interest for developing navigation solutions for autonomous agents in adversarial environments, such as delivery vehicles or vehicles operating in war zones. This paper develops a theory of continuous ambush games for environments with obstacles. It draws a parallel between discrete and continuous ambush games through topological analysis for maximal flow computation. Two complementary approaches are adopted.

First, we recall some concepts on ambush games on graphs with vertex flow capacity constraints. The optimal solution of the game is proven to be a uniform probability distribution over a set of $\kappa$ fully separated paths on the graph $G$, where $\kappa$ is the minimal vertex cut of $G$.

Second, a continuous approach, requiring the simplifying but realistic assumption that the environment is polygonal, allows us to compute theoretical optimal solutions. The optimal value of a continuous ambush game, as defined in Section III, is $\frac{1}{l_R(\Omega)^*}$, where $R$ measures the reach of the adversary and $l_R(\Omega)^*$ represents the maximum number of paths from $s$ to $t$ separated by at least $2R$ at any point, or $2R$-WSP. The optimal strategies for both agents depend solely on the geometry of the environment and on the reach of the adversary.

Third, the problem of ambush games in a continuous environment is tackled by sampling the environment and creating a linear problem based on the resulting graph, as presented in Section IV. Numerical examples indicate that the optimal strategy returned by the linear program converges towards the theoretical optimal strategy as the number of vertices in the graph representation of the environment increases. A proof is provided, that confirm this convergence for graphs created using the Rapidly exploring Random Graph algorithm.

Of particular interest to us are the two solutions to the continuous problem, which both exhibit interesting features yet have their own limitations. On the one hand, the purely continuous approach enables a truly continuous comprehension and modeling of the environment studied. Assuming a polygonal environment, a strong yet realistic assumption, theoretical proofs regarding the optimality of the solution are derived. However, the corresponding algorithms are for now limited to two dimensional environments [17], and extensions to higher dimensions might prove difficult.

On the other hand, the sampling approach only requires an admissibility predicate [27], meaning whether a point of the environment is collision-free with respect to the obstacles and feasible with respect to the constraints. This approach allows for the creation of a graph on which a solution of the ambush game can be obtained by running optimization problems as described in Section IV. The upside of this approach is its simple implementation, similar to sampling-based path planning approaches, which allows to build a lattice or a Rapidly-Exploring Dense Graph ([28],[29]) and then run the optimization on the resulting network. Moreover, this approach can easily be generalized to higher dimensions, which could include velocity features for instance, allowing for non uniform reward functions depending on the velocity of the agent. A kinodynamic extension of the problem is envisioned in future work. The state space will take into account position and velocity. This work can also be adapted to include a vehicle model and its performance, to restrain the network to only a set of feasible maneuvers for BLUE.

## Acknowledgements

This work was supported by the Army Research Office under MURI Award W911NF-11-1-0046.

## VI. Appendix

### A. Vertex capacity constrained network flow theory

In this section, we detail the proofs of the extension of some results of network flow theory to networks with flow capacity constrained vertices. We begin by providing an upperbound on the maximum flow on such a network.

*Proof. of Proposition II.1*
Let $[S, C, \overline{S}]$ be any vertex cut with $s \in S$ and $t \in \overline{S}$. We have,

$$\begin{aligned} val\ f =& \sum_{v\in S} f^+(v) - \sum_{v\in S} f^-(v),\\ =& \sum_{\substack{v\in S, u\in\overline{S},\\ (v,u)\in E}} f_{vu} + \sum_{\substack{v\in S, u\in S,\\ (v,u)\in E}} f_{vu} + \sum_{\substack{v\in S, u\in C,\\ (v,u)\in E}} f_{vu} \\ &- \sum_{\substack{v\in S, u\in\overline{S},\\ (u,v)\in E}} f_{uv} - \sum_{\substack{v\in S, u\in S,\\ (u,v)\in E}} f_{uv} - \sum_{\substack{v\in S, u\in C,\\ (u,v)\in E}} f_{uv}. \end{aligned}$$

But,

$$\sum_{\substack{v\in S, u\in S,\\ (v,u)\in E}} f_{vu} = \sum_{\substack{v\in S, u\in S,\\ (u,v)\in E}} f_{uv}.$$

And, by definition of $C$, there are no arcs between $S$ and $\overline{S}$:

$$\begin{cases} \displaystyle\sum_{\substack{v\in S, u\in\overline{S},\\ (v,u)\in E}} f_{vu} = 0,\\ \displaystyle\sum_{\substack{v\in S, u\in\overline{S},\\ (u,v)\in E}} f_{uv} = 0. \end{cases}$$

Hence

$$\begin{aligned} val\ f =& \sum_{\substack{v\in S, u\in C,\\ (v,u)\in E}} f_{vu} - \sum_{\substack{v\in S, u\in C,\\ (u,v)\in E}} f_{uv},\\ =& c([S, C, \overline{S}]) - \sum_{\substack{v\in S, u\in C,\\ (u,v)\in E}} f_{uv}. \end{aligned}$$

Recall that the flow on each arc is a non-negative value, therefore

$$\sum_{\substack{v\in S, u\in C,\\ (u,v)\in E}} f_{uv} \geq 0.$$

And, finally,

$$val\ f \leq c([S, C, \overline{S}]). \tag{8}$$

□

This results allows for the extension of Ford-Fulkerson Theorem to the type of networks studied.

*Proof. of Theorem II.1*
Assume that the network does not include any isolated vertex, then each vertex has at least one incoming arc of capacity one. Therefore an arc is saturated only if both its incoming and outgoing vertices are saturated. However the converse is not true: for example if a given vertex has two incoming arcs and two outgoing arcs with flows of capacity equal to 1/2, then the vertex is saturated while none of the arcs is saturated.

Proposition II.1 tells us that the value of any flow $f$ is less than or equal to the capacity of any cut $K = [S, C, \overline{S}]$ in $G$. Therefore if $f$ is a flow and $K$ a cut of $G$ such that $val\ f = cap\ K$, then $f$ is a maximum flow and $K$ is a minimum cut. Reproducing the proof of Ford-Fulkerson theorem ([18], p.64) for constrained flow at vertices, let $F$ be a maximum flow in $G$ with $val\ f = w_0$ and $S$ be a subset of $G$ built recursively, such that:

- $s \in S$, and

- If a vertex $u \in S$, and either $(u,v) \in E$ and $f_v < c(v)$ or $f_{vu} > 0$, then include $v$ in $S$.

And the cut can be defined as follows,

- $C = \{v \in G \setminus S | u \in S, (u,v) \in E\}$,
- $\overline{S} = G \setminus (S \cup C)$.

By definition of $K$, $t$ does not belong to $C$. We claim that $t$ cannot belong to $S$. Suppose $t \in S$. Then there exists a path $P$ from $s$ to $t$, say $P = sv_1v_2 \cdots v_kt$, with vertices in $S$ such that for any arc of $P$, either $f_{v_{j+1}} < c(v_{j+1})$ or $f_{v_{j+1}v_j} > 0$. Call an arc joining $v_j$ and $v_{j+1}$ of $P$ a *forward arc* if it is directed from $v_j$ to $v_{j+1}$; otherwise, it is a *backward arc*. Let $\delta_1$ be the minimum of all differences $c(v_{j+1}) - f(v_{j+1})$ for forward arcs, and let $\delta_2$ be the minimum of all flows in backward arcs of $P$. Both $\delta_1$ and $\delta_2$ are positive, by the definition of $S$. Let $\delta = \min\{\delta_1, \delta_2\}$. Increase the flow in each forward arc of $P$ by $\delta$ and also decrease the flow in each backward arc of $P$ by $\delta$. Keep the flows along the other arcs of $G$ unaltered. Then there results a new flow whose value is $w_0 + \delta > w_0$, leading to a contradiction. [This is because among all arcs incident at $s$, only in the initial arc of $P$, the flow value is increased by $\delta$ if it is a forward arc or decreased by $\delta$ if it is a backward arc.] This contradiction shows that $t \notin S$ and therefore $t \in \overline{S}$.

Therefore $[S, C, \overline{S}]$ is a cut separating $s$ and $t$. In particular, if $v \in S$ and $u \in \overline{S}$, we have, by the definition of $S$, $f_{vu} = c(v,u)$ if $(v,u)$ is an arc of $G$, and $f_{uv} = 0$ if $(u,v)$ is an arc of $G$. Hence,

$$w_0 = \sum_{\substack{u \in S, \\ v \in \overline{S}}} f_{uv} - \sum_{\substack{u \in S, \\ v \in \overline{S}}} f_{vu} = \sum_{\substack{u \in S, \\ v \in \overline{S}}} f_{uv} - 0 = c([S, C, \overline{S}]). \quad (9)$$

□

Finally, this leads to the vertex capacity constrained version of Menger's Theorem, which relates the number of distinct $(s-t)$ paths to the the maximum flow of such a network.

*Proof.* *of Theorem II.2*
Let $f^*$ be a maximum flow in $G$ and let $\{f_i\}_{k=1}^{\kappa}$ be any system of $\kappa$ vertex-disjoint directed ($s$-$t$)-paths in $G$. Define a function $f$ on $E$ by

$$f(a) = \begin{cases} 1 & \text{if } (i,j) \in f_k \text{ for some } k, \\ 0 & \text{otherwise.} \end{cases}$$

Then $f$ is a flow in $G$ with value $\kappa$. Since $f^*$ is a maximum flow, we have $val\ f^* \geq \kappa$. Now, assume that there are strictly fewer disjoint paths than the maximum flow. Since the flow capacity of each vertex is 1, there are at least $k = \lceil val\ f^* \rceil \geq \kappa + 1$ distinct paths. But, by definition of $\kappa$, we have $k \leq \kappa$. This is a contradiction, therefore $\kappa \geq val\ f^*$. □

### B. Necessary condition for optimality of RED's strategy

The following proposition shows that if RED's strategy is not optimized on a subset of $V$ that contains a min-cut of $G$, then BLUE wins all the time since there is a walk from $s$ to $t$ that is not covered by RED.

**Proposition VI.1.** *If RED's strategy space support $Q \subset V$ does not contain a vertex cut of $G$, then $\mathcal{V}_{EQ} = \min_{p \in E} \max_{q \in Q} p^T D q = 0$.*

*Proof.* If $Q$ does not contain an $s$-$t$ cut then there exists $G' = (V', E') \subset G$ such that:

$$\begin{cases} s, t \in V', \\ V' \cap Q = \emptyset, \\ G' \text{ is connected.} \end{cases}$$

Therefore a walk $w = (V_w, E_w)$ from $s$ to $t$ can be constructed such that there is no ambush on $w$: $\forall n \in V_w, q_n = 0$. Assigning a probability of 1 to this walk results in an upper-bound $\overline{\mathcal{V}}$ of the optimal value $\mathcal{V}_{PQ}$ of the reduced game on $P \times Q$ since only a subset of BLUE's strategy space is used and BLUE is the minimizer. This is formulated as follows:

$$p_{ij} = 0, \forall (i,j) \in E' \setminus E_w,$$
$$q_j = 0, \forall j \in V \setminus Q.$$

$$\begin{aligned} \overline{\mathcal{V}} &= \sum_{j \in V} \sum_{i | (i,j) \in E'} p_{ij} q_j \\ &= \sum_{j \in Q} \sum_{i | (i,j) \in E'} p_{ij} q_j + \sum_{j \in V \setminus Q} \sum_{i | (i,j) \in E'} p_{ij} q_j \\ &= \sum_{j \in Q} \sum_{i | (i,j) \in E'} p_{ij} q_j \\ &= \sum_{j \in Q} \sum_{i | (i,j) \in E' \setminus E_w} p_{ij} q_j + \sum_{j \in Q} \sum_{i | (i,j) \in E_w} p_{ij} q_j \\ &= \sum_{j \in Q} \sum_{i | (i,j) \in E_w} p_{ij} q_j \\ &= 0. \end{aligned}$$

Since $V_w \subset V'$ and $Q \cap V' = \emptyset$ therefore if $j \in Q$ then $\{i | (i,j) \in E_w\} = \emptyset$. Now $\mathcal{V}_{EQ}$ is a sum of products of non-negative values therefore

$$0 \leq \mathcal{V}_{EQ} \leq \overline{\mathcal{V}} = 0.$$

This property is illustrated in Figure 11. □

### C. Probabilistic optimality and completeness of Sampled Ambush Game

In this section, we recall some concepts from the sampling based path planning literature and algorithms introduced by Karaman and Frazzoli ([23]) in order to prove Theorem IV.1.

Let $\chi = (0,1)^d$ be the configuration space, where $d \in \mathbb{N}$, $d \geq 2$. Let $\chi_{\text{obs}}$ be the obstacle region, such that $\chi \setminus \chi_{\text{obs}}$ is an open set, and denote the obstacle-free space as $\chi_{\text{free}} = cl(\chi \setminus \chi_{\text{obs}})$, where $cl()$ denotes the closure of a set. The initial condition $x_{\text{init}}$ is an element of $\chi_{\text{free}}$, and the goal region $\chi_{\text{goal}}$ is an open subset of $\chi_{\text{free}}$. A path planning problem is defined by a triplet $(\chi_{\text{free}}, x_{\text{init}}, \chi_{\text{goal}})$.

**Definition VI.1.** *(Total variation)* Let $\sigma : [0,1] \to \mathbb{R}^d$; the total variation of $\sigma$ is defined as

$$TV(\sigma) = \sup_{n \in \mathbb{N}, 0 = \tau_0 < \tau_1 < \cdots < \tau_n = 1} \sum_{i=1}^{n} |\sigma(\tau_i) - \sigma(\tau_{i-1})|.$$

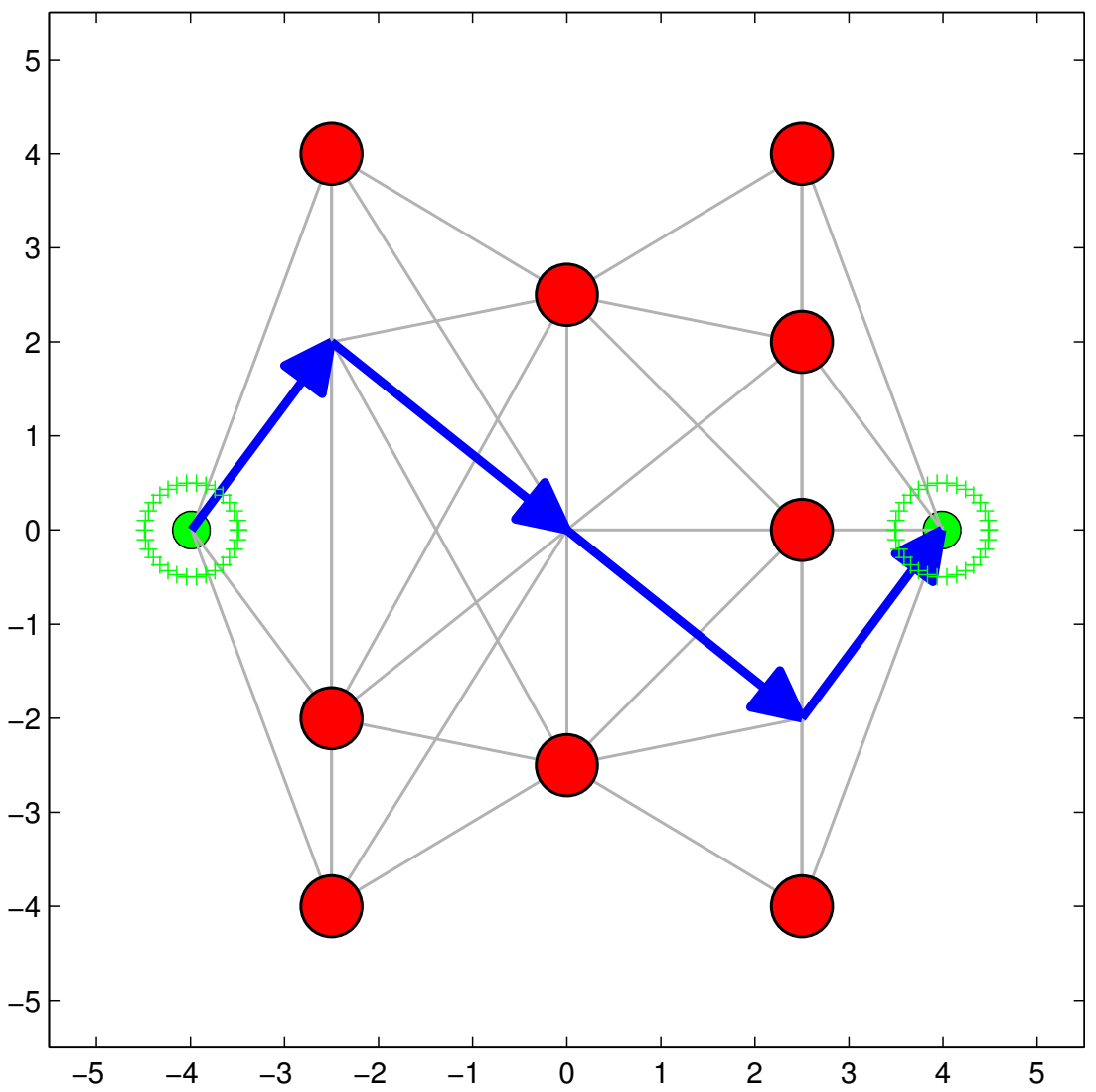

Fig. 11. (Proposition VI.1) RED's accessible space $Q$ is represented by the RED circle. BLUE's path, represented by the BLUE arrows, does not intersect with RED's accessible space, therefore $\mathcal{V}_{\mathcal{E}\mathcal{Q}} = 0$. The green areas around origin and destination illustrate the fact that the local outcome is zero at these locations.

A function with $TV(\sigma) < \infty$ is said to exhibit *bounded variation*.

**Definition VI.2.** *(Path)* A continuous function $\sigma : [0,1] \to \mathbb{R}^d$ of bounded variation is called a path.

The total variation of a path is essentially its length, i.e. the Euclidean distance traversed by the path in $\mathbb{R}^d$.

Following the intuition behind the Sampled Continuous ambush Game (SCAG), we are interested in sampling based path planning algorithms whose outputs are graphs with vertices and edges representing points and paths in $\chi_{\text{free}}$. Examples of such algorithms are PRM* and RRG [23], which are detailed in Algorithms 1 and 2.

**Algorithm 1** PRM*

$V \leftarrow \{x_{\text{init}} \cup \{\text{SampleFree}_i\}_{i=1,\cdots,n}; E \leftarrow \emptyset$
**foreach** $v \in V$ **do**:
  $U \leftarrow \text{Near}\big(G = (V,E), v, \gamma_{\text{PRM}}(\log(n)/n)^{1/d}\big) \setminus \{v\}$;
  **foreach** $u \in U$ **do:**
    **if** CollisionFree$(v,u)$ **then**:
      $E \leftarrow E \cup \{(v,u),(u,v)\}$
**Return** $G = (V,E)$.

where

- the function Near : $(G,x,r) \mapsto V' \subseteq V$ returns the vertices in $V$ that are contained in a ball of radius $r$ centered at $x$, i.e.

$$\text{Near}(G=(V,E),x,r) := \{v \in V : v \in \mathcal{B}_{x,r}\}.$$

- the function Nearest : $(G,x) \mapsto v \in V$ returns the vertex in $V$ that is "closest" to $x$ in terms of a given distance function.

$$\text{Nearest}(G=(V,E),x) := \arg\min \|xv\|.$$

**Algorithm 2** RRG

$V \leftarrow \{x_{\text{init}}\}; E \leftarrow \emptyset$
**for** $i = 1,\cdots,n$ **do**:
$x_{\text{rand}} \leftarrow \text{SampleFree}_i$;
$x_{\text{nearest}} \leftarrow \text{Nearest}(G=(V,E),x_{\text{rand}})$;
$x_{\text{new}} \leftarrow \text{Steer}(x_{\text{nearest}},x_{\text{rand}})$;
**if** ObstacleFree$(x_{\text{nearest}},x_{\text{rand}})$ **then**:
  $X_{\text{near}} \leftarrow \text{Near}\big(G = (V,E), x_{\text{new}}, \min\{\gamma_{\text{RRG}}(\log(\text{card}(V))/\text{card}(V))^{1/d},\eta\}\big)$;
  $V \leftarrow V \cup \{x_{\text{new}}\}$;
  $E \leftarrow E \cup \{(x_{\text{nearest}},x_{\text{new}}),(x_{\text{new}},x_{\text{nearest}})\}$
  **foreach** $x_{\text{near}} \in X_{\text{near}}$ **do:**
    **if** CollisionFree$(x_{\text{near}},x_{\text{new}})$ **then**:
      $E \leftarrow E \cup \{(x_{\text{near}},x_{\text{new}}),(x_{\text{new}},x_{\text{near}})\}$
**Return** $G = (V,E)$.

- the function Steer : $(x,y) \mapsto z$ returns a point $z \in X$ such that $z$ is "closer" to $y$ than $x$ is:

$$\text{Steer}(x,y) := \arg\min_{z \in \mathcal{B}_{x,\eta}} \|zy\|.$$

When analyzing sampling based path planning algorithms, attention is given to the existence and optimality of the solutions returned by these algorithms as the number of points in the graph increases towards infinity. If the vertex sampling is dense, one can prove that some algorithms find a solution almost surely as the number of samples increases. These algorithms are said to be *probabilistically complete*. A subclass of the class of probabilistically complete algorithms has an even more interesting property, which is that they identify the optimal solution of the optimization problem almost surely as the number of vertices goes to infinity. These algorithms are said to be *asymptotically optimal*. Algorithms 1 and 2 are examples of such algorithms.

**Definition VI.3.** *(Probabilistic completeness)* An algorithm ALG is probabilistically complete, if, for any robustly feasible path planning problem $(\chi_{\text{free}}, x_{\text{init}}, \chi_{\text{goal}})$,

$$\liminf_{n\to\infty} \mathbb{P}(\{\exists x_{\text{goal}} \in V_n^{ALG} \cap \chi_{\text{goal}} \text{ such that } x_{\text{init}} \text{ is connected to } x_{\text{goal}} \text{ in } G_n^{ALG}\}) = 1,$$

where $G_n^{ALG} = (V_n^{ALG}, E_n^{ALG})$ is the graph obtained after $n$ iterations of the $ALG$ algorithm, and $\mathbb{P}(A)$ is the probability that event $A$ occurs.

**Definition VI.4.** *(Asymptotic optimality)* Let $Y_n^{\text{ALG}}$ be the extended random variable corresponding to the cost of the minimum-cost solution included in the graph returned by ALG at the end of iteration $n$. A path planning algorithm ALG is asymptotically optimal if, for any path planning problem $(\chi_{\text{free}}, x_{\text{init}}, \chi_{\text{goal}})$ and cost function $c : \Sigma \to \mathbb{R}_{\geq 0}$ that admit a robustly optimal solution with finite cost $c^*$,

$$\mathbb{P}\big(\{\limsup_{n\to\infty} Y_n^{\text{ALG}} = c^*\}\big) = 1.$$

In order to prove asymptotic optimality of PRM* and RRG, an additional hypothesis is required, which describes that the optimal solution to the path planning problem should be homotopically equivalent to a feasible path with strong $\delta$-clearance, as defined below.

**Definition VI.5.** *(Strong $\delta$-clearance)[23]* Let $\delta > 0$ be a real number. A state $x \in \chi_{\text{free}}$ is said to be a $\delta$-interior state of $\chi_{\text{free}}$, if the closed ball of radius $\delta$ centered at $x$ lies entirely inside $\chi_{\text{free}}$. The $\delta$-interior of $\chi_{\text{free}}$, denoted as $\text{int}_\delta(\chi_{\text{free}})$, is defined as the collection of all $\delta$-interior states, i.e. $\text{int}_\delta(\chi_{\text{free}}) := \{x \in \chi_{\text{free}} | \mathcal{B}_{x,\delta} \in \chi_{\text{free}}\}$. In other words, the $\delta$-interior of $\chi_{\text{free}}$ is the set of all states that are at least a distance $\delta$ away from any point in the obstacle set. A collision- free path $: [0,1] \to \chi_{\text{free}}$ is said to have strong $\delta$- clearance, if $\sigma$ lies entirely inside the $\delta$-interior of $\chi_{\text{free}}$, i.e. $\sigma(\tau) \in \text{int}_\delta(\chi_{\text{free}})$ for all $\tau \in [0,1]$. A path planning problem $(\chi_{\text{free}}, x_{\text{init}}, \chi_{\text{goal}})$ is said to be robustly feasible if there exists a path with strong $\delta$-clearance, for some $\delta > 0$.

**Definition VI.6.** *(Weak $\delta$-clearance)[23]* A collision-free path $\sigma : [0,1] \to \chi_{\text{free}}$ is said to have weak $\delta$-clearance, if there exists a path $\sigma'$ that has strong $\delta$-clearance and there exists a homotopy $\psi$, with $\psi(0) = \sigma$, $\psi(1) = \sigma'$, and for all $\alpha \in (0,1]$ there exists $\delta_\alpha > 0$ such that $\psi(\alpha)$ has strong $\delta_\alpha$-clearance.

Assuming that there exists $\delta > 0$ such that the optimal solution to the path planning problem has weak $\delta$-clearance, Karaman and Frazzoli proved asymptotic optimality of both RRG and PRM$^*$.

**Theorem VI.1.** *(Asymptotic optimality of PRM$^*$)[23]*
*If $\gamma_{PRM} > 2(1 + 1/d)^{1/d}\big(\frac{\mu(\chi_{free})}{\zeta_d}\big)^{1/d}$, then the PRM$^*$ algorithm is asymptotically optimal.*

**Theorem VI.2.** *(Asymptotic optimality of RRG)[23]*
*If $\gamma_{RRG} > 2(1 + 1/d)^{1/d}\big(\frac{\mu(\chi_{free})}{\zeta_d}\big)^{1/d}$, then the RRG algorithm is asymptotically optimal.*

Using the concepts previously introduced, a proof of Theorem IV.1 can now be detailed.

*Proof. (Theorem IV.1)* Consider a polygonal environment $\Omega$. Let $\kappa \in \mathbb{N}$ be the maximum number of $2R$ WSP in $\Omega$ and let $\Sigma = \{\sigma^k\}_{k=1}^{\kappa}$ be such a set of $2R$-WSP. Assume that there exists a set $\{\delta^k\}_{k=1}^{\kappa}$ such that for all $k = 1, \cdots, \kappa$, $\delta^k > 0$ and $\sigma^k$ has weak $\delta^k$ clearance.

Let $k \in [1, \kappa]$. Let $\{\delta_n^k\}_{n\in\mathbb{N}}$ be a real sequence such that $\delta_n^k > 0$ for all $n \in \mathbb{N}$ and $\delta_n^k$ approaches $\delta^k$ as $n$ approaches infinity. Given that PRM$^*$ and RRG share similar properties, the denomination ALG will be used in the remainder of this proof to denote either algorithm. Consider the sampled graph $G_n^{\text{ALG}} = (V_n^{\text{ALG}}, E_n^{\text{ALG}})$ output from ALG after $n$ points have been sampled. Karaman and Frazzoli [23] show that a sequence $\{\sigma_n^k\}_{n\in\mathbb{N}}$ of paths can be constructed such that $\sigma_n^k$ has strong $\delta_n^k$-clearance for all $n \in \mathbb{N}$, and $\sigma_n^k$ converges to $\sigma^k$ as $n$ approaches infinity.

Let $P_n$ denote the set of all paths in the graph. Let $\sigma_n^{k'}$ be the path that is closest to $\sigma_n^k$ in terms of bounded variation norm among all paths in $P_n$, i.e.

$$\sigma_n^{k'} = \arg\min_{\sigma\in P_n} \|\sigma - \sigma_n^k\|_{\text{BV}}.$$

Then, by Lemma 55 of [23], the random variable $\|\sigma_n^{k'} - \sigma_n^k\|_{\text{BV}}$ converges to zero almost surely, i.e

$$\mathbb{P}\big(\{\lim_{n\to\infty} \|\sigma_n^{k'} - \sigma_n^k\|_{\text{BV}} = 0\}\big) = 1.$$

And,

$$\mathbb{P}\big(\{\lim_{n\to\infty} \|\sigma_n^{k'} - \sigma^k\|_{\text{BV}} = 0\}\big) = 1,$$

where $\|\sigma\|_{\text{BV}} = \int_0^1 \|\sigma(\tau)\| d\tau + TV(\sigma)$.

Consider now the set of all paths in $\Sigma$. The optimal outcome of the Sampled Continuous Ambush Game (SCAG) is equal to the outcome of the CAG if $\kappa$ $2R$-WSP can be found on $G_n^{\text{ALG}}$. Using the random sequences $\{\sigma_n^{k'}\}_{n\in\mathbb{N},k=1,\cdots,\kappa}$ defined above, we can define the event that this occurs as the event where, for any pair of $2R$-WSP in $\Sigma$, the corresponding approximate paths on $G_n^{\text{ALG}}$ are also $2R$-WSP. Let $A_n$ characterize this event after $n$ iterations of ALG:

$$A_n = \bigvee_{j,k=1,..,\kappa,\ j\neq k} \Big\{ \inf_{s,t\in[0,1]} \|\sigma_n^{j'}(s) - \sigma_n^{k'}(t)\| > 2R \ \Big|\ \inf_{s,t\in[0,1]} \|\sigma^j(s) - \sigma^k(t)\| > 2R \Big\}$$

By definition, for all $j, k \in 1, \cdots, \kappa$, for all $s, t \in [0,1]$, we have $\|\sigma^j(s) - \sigma^k(t)\| > 2R$. Define $\epsilon > 0$ such that

$$\epsilon = \min_{j,k\in 1,\cdots,\kappa, s,t\in[0,1]} \|\sigma^j(s) - \sigma^k(t)\| - 2R.$$

Let $N_j$ be such that for all $n$ greater than $N_j$, we have

$$\|\sigma^{j'} - \sigma^j\|_{\text{BV}} < \frac{\epsilon}{2}.$$

Then,

$$\forall n > N_j,\ \forall s, t \in [0,1], \qquad \|\sigma_n^{j'}(s) - \sigma_n^j(t)\| < \frac{\epsilon}{2}.$$

Define $N$ as the maximum element of the sequence $\{N_j\}_{j=1,\cdots,\kappa}$, and let $n \in \mathbb{N}$ be greater than $N$. Let $j$ and $k$ be two different integers in $\{1, \cdots, \kappa\}$. Then for all $s, t \in [0,1]$,

$$\begin{aligned}
\|\sigma_n^{j'}(s) - \sigma_n^{k'}(t)\| &= \|\sigma_n^{j'}(s) - \sigma_n^j(t) + \sigma_n^j(t) - \sigma_n^{k'}(t)\| \\
&\geq |\ \|\sigma_n^j(t) - \sigma_n^{k'}(t)\| - \|\sigma_n^{j'}(t) - \sigma_n^j(s)\|\ | \\
&\geq \|\sigma_n^j(t) - \sigma_n^{k'}(t)\| - \frac{\epsilon}{2} \\
&\geq |\ \|\sigma_n^j(t) - \sigma_n^k(s)\| - \|\sigma_n^k(s) - \sigma_n^{k'}(t)\|\ | - \frac{\epsilon}{2} \\
&\geq \|\sigma_n^j(t) - \sigma_n^k(s)\| - \epsilon \\
&> 2R.
\end{aligned}$$

Therefore,

$$\mathbb{P}\big(\|\sigma_n^{j'}(s) - \sigma_n^{k'}(t)\| > 2R \ \mid\ \sigma^j(s) - \sigma^k(t)\| > 2R\big) = 1,\ \forall n > N,$$

and,

$$\mathbb{P}(A_n) = 1, \qquad \forall n > N.$$

Finally,

$$\mathbb{P}\big(\lim_{n\to\infty} A_n\big) = 1.$$

Let $\mathcal{V}_n^{\text{ALG}}$ be the outcome of SCAG after $n$ iterations of ALG. The fact that the event $A_n$ happens with probability one means that $G_n^{\text{ALG}}$ contains at least $\kappa$ $2R$-WSP, therefore

$$\mathcal{V}_n^{\text{ALG}} \leq \frac{1}{\kappa}.$$

Now, the optimal outcome of the SCAG is lower bounded by the optimal outcome of the CAG. By assumption, the environment contains at most $\kappa$ $2R$-WSP, and

$$\mathcal{V}_n^{\text{ALG}} \geq \mathcal{V}_{\text{CAG}}^* = \frac{1}{\kappa}, \qquad \forall n > N.$$

We have shown that for all $n$ greater than $N$,

$$\mathcal{V}_n^{\text{ALG}} = \frac{1}{\kappa}.$$

$$\mathbb{P}\big(\lim_{n\to\infty} \mathcal{V}_n^{\text{ALG}} = \frac{1}{\kappa}\big) = 1.$$

□

Note that Theorem IV.1 proves a necessary condition for ensuring convergence of the SCAG to the optimal outcome of the CAG. If the graph creation algorithm allows to asymptotically approximate any path as the number of samples increases to infinity, then the outcome of the SCAG converges to the optimal outcome of the CAG almost surely. However, there might be graph creation algorithms that are not asymptotically optimal in the path planning sense but still allow to create at optimal set of $2R$-WSP. We suspect, for example, that convergence occurs even for an 8-connected grid, where the total variation of the created paths may remain different from that of the corresponding well-separated paths, yet it is possible to create an 8-connected path "sufficiently close" to these paths.